\pdfoutput=1
\documentclass[11pt]{article}

\usepackage[final]{acl}

\usepackage{times}
\usepackage{setspace} % 引入 setspace 包
\usepackage{latexsym}
\usepackage{tabularray}  % for toprule, midrule, bottomrule
\usepackage{multirow} % for multirow
\usepackage{float}
\usepackage{amssymb}
\usepackage{booktabs}
\usepackage{algorithmic}
\usepackage{algorithm}
\usepackage{amsmath}
\usepackage{longtable}
\usepackage{listings}
\usepackage{xcolor}
\usepackage[T1]{fontenc}
\usepackage[utf8]{inputenc}
\usepackage{microtype}
\usepackage{inconsolata}

\usepackage{graphicx}
\definecolor{MineShaft}{rgb}{0.2,0.2,0.2}

\title{Error Classification of Large Language Models on Math Word Problems: A Dynamically Adaptive Framework}

\author{
Yuhong Sun\textsuperscript{$\heartsuit$}\thanks{Equal contribution.} \quad 
Zhangyue Yin\textsuperscript{$\diamondsuit$}$^*$ \quad 
Xuanjing Huang\textsuperscript{$\diamondsuit$} \quad 
Xipeng Qiu\textsuperscript{$\diamondsuit$}\textsuperscript{\dag} \quad
Hui Zhao\textsuperscript{$\heartsuit$ $\spadesuit$}\textsuperscript{\dag} \\
\textsuperscript{$\heartsuit$}Software Engineering Institute, East China Normal University\\
\textsuperscript{$\diamondsuit$}School of Computer Science and Artificial Intelligence, Fudan University \\
\textsuperscript{$\spadesuit$}Shanghai Key Laboratory of Trustworthy Computing, Shanghai, China \\
\texttt{sunyuhong@stu.ecnu.edu.cn}  \quad
\texttt{yinzy21@m.fudan.edu.cn}\\
\texttt{\{xjhuang,xpqiu\}@fudan.edu.cn} \quad
\texttt{hzhao@sei.ecnu.edu.cn}
}

\begin{document}
\maketitle

\begin{abstract}
Large Language Models (LLMs) have demonstrated remarkable capabilities across various domains. 
Math Word Problems (MWPs) serve as a crucial benchmark for evaluating LLMs' reasoning abilities. 
While most research primarily focuses on improving accuracy, it often neglects understanding and addressing the underlying patterns of errors.
Current error classification methods rely on static and predefined categories, which limit their ability to capture the full spectrum of error patterns in mathematical reasoning.
To enable systematic error analysis, we collect error samples from 15 different LLMs of varying sizes across four distinct MWP datasets using multiple sampling strategies.
Based on this extensive collection, we introduce MWPES-300K, a comprehensive dataset containing 304,865 error samples that cover diverse error patterns and reasoning paths.
To reduce human bias and enable fine-grained analysis of error patterns, we propose a novel framework for automated dynamic error classification in mathematical reasoning.
Experimental results demonstrate that dataset characteristics significantly shape error patterns, which evolve from basic to complex manifestations as model capabilities increase.
With deeper insights into error patterns, we propose Error-Aware Prompting (EAP) that incorporates common error patterns as explicit guidance, leading to significant improvements in mathematical reasoning performance.
\end{abstract}

\begin{figure}[t]
  \centering
  \includegraphics[width=0.4\textwidth]{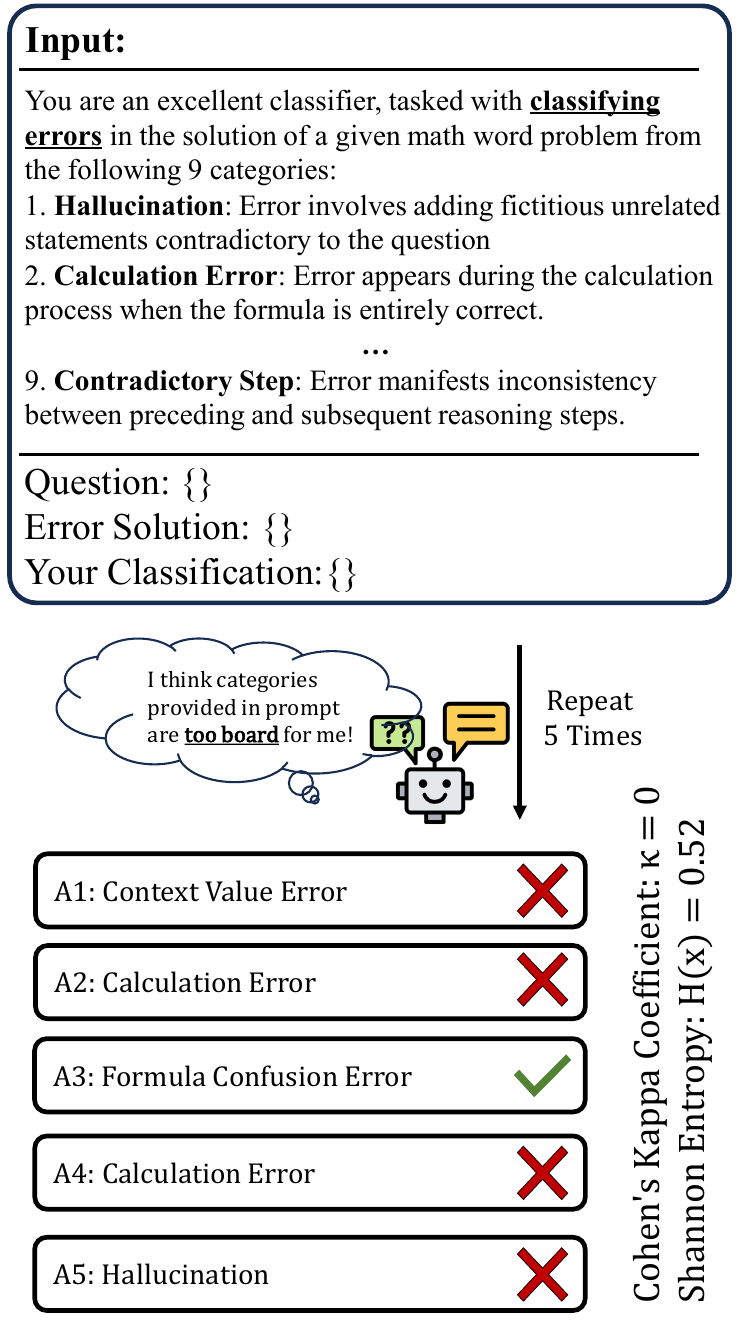}
  \caption{LLMs struggle to categorize errors into predefined, broad and ambiguous categories, resulting in low consistency and accuracy.}
  \label{fig:sampling}
  \vspace{-1.5em}
\end{figure}

\section{Introduction}
\label{sec:intro}

Large Language Models (LLMs) have showcased remarkable capabilities across various domains, yet mathematical reasoning remains a notable challenge~\citep[][\emph{inter alia}]{brown2020languagemodelsfewshotlearners, achiam2023gpt, team2023gemini, dubey2024llama, jaech2024openai}. Mathematical Word Problems (MWPs) serve as a critical benchmark for evaluating LLMs' mathematical reasoning abilities, as they demand both natural language understanding and mathematical computation skills. Recent advancements, such as Chain-of-Thought (CoT) prompting~\citep{wei2022chain} and specialized mathematical training~\citep{qwen25math, deepseekmath}, have shown promising progress in this domain. However, these improvements primarily focus on accuracy metrics while overlooking the critical aspect of understanding and addressing the underlying patterns of errors.

The analysis of LLM errors in mathematical reasoning has emerged as a vital research direction. \citet{lightman2024lets} introduces PRM800K, a dataset featuring manually annotated solution steps to enhance model performance through pre-training. Building on this foundation, \citet{zeng2024mrgsm8kmetareasoningbenchmarklarge} develops MR-GSM8K, which incorporates potentially erroneous reasoning paths and requires models to both evaluate solution correctness and pinpoint error locations. This approach aligns with \citet{EIC}'s work, which requires models to not only identify but also correct errors.

However, current approaches to error analysis in mathematical reasoning face several fundamental deficiencies. Traditional error classification methods rely heavily on static, predefined categories that are based on human observation and empirical judgment. This approach proves inadequate for capturing the diverse and evolving nature of LLM errors, particularly when scaling to large datasets. The challenge is further complicated by the observation that different LLMs often generate distinct types of errors when solving the same problem, making static classification frameworks increasingly obsolete. For instance, GPT-3.5 and LLaMA-3.1-70B may produce entirely different types of errors when solving the same MWP, necessitating a more flexible classification approach.

Moreover, as illustrated in Figure~\ref{fig:sampling}, predefined error taxonomies often employ broad, ambiguous categories such as \textit{"Hallucination"} or \textit{"Calculation Error"} that lack the granularity needed for meaningful analysis. This oversimplification obscures the root causes of errors and hampers the development of targeted improvements. Additionally, current methodologies typically focus solely on identifying the first error in a solution path, overlooking the potential cascade of subsequent errors that could provide valuable insights into model behavior.

To facilitate systematic error analysis, we introduce MWPES-300K (Math Word Problem Error Solutions), a comprehensive dataset containing 304,865 erroneous MWP solutions. Unlike existing error analysis datasets that focus on single dataset or synthetic errors, MWPES-300K captures real error patterns by collecting model outputs across four diverse MWP datasets.  Furthermore, we propose a novel framework for automated dynamic error classification in mathematical reasoning. Our approach adaptively evolves error categories based on observed model outputs, enabling fine-grained analysis of error patterns while reducing human bias and intervention. Moreover, we introduce an innovative Error-Aware Prompting (EAP) mechanism that explicitly guides models to avoid potential errors during problem-solving, substantially enhancing mathematical reasoning performance. Our main contributions are as follows:

\begin{itemize}
    \item Development of the first automated framework for dynamic error classification in MWPs, adapting to diverse error patterns across different LLMs and problem types.
    
    \item Introduction of MWPES-300K, a comprehensive dataset containing 304,865 error samples collected from 15 different LLMs across 4 MWP datasets of varying difficulty levels using multiple sampling strategy, enabling robust analysis of error patterns and distribution.
    
    \item Implementation of an Error-Aware Prompting (EAP) mechanism that significantly improves mathematical reasoning performance by explicitly guiding models to avoid relevant error patterns through our framework.
\end{itemize}

\begin{table*}
\centering
\begin{tblr}{
  width = \linewidth,
  colspec = {Q[253]Q[162]Q[135]Q[150]Q[108]Q[141]},
  column{even} = {c},
  column{3} = {c},
  column{5} = {c},
  hline{1,8} = {-}{0.08em},
  hline{2} = {-}{0.05em},
  hline{7} = {-}{0.08em}
}
\textbf{Dataset}          & \textbf{Dataset Size}           & \textbf{Injected Errors} & \textbf{Error Categories} & \textbf{Models Covered} & \textbf{Dataset Count} \\
PRM800K          & 98,732 & \textbf{×}      & 7                & 1          & 1                  \\
MathCritique-76k & 76,000                 & \checkmark      & Unspecified                & 1          & 2                  \\
MathCheck-GSM    & 516                    & \checkmark      & 4                & 1          & 1                  \\
MR-GSM8K         & 3,000                  & \checkmark      & Unspecified                & 1          & 1                  \\
EIC-Math         & 1,800                  & \checkmark      & 9                & 1          & 2                  \\
\textbf{MWPES-300k(ours)}       & \textbf{304,865}                & \textbf{×}      & \textbf{Dynamic}          & \textbf{15}         & \textbf{4}                  
\end{tblr}
\caption{Comparison of MWP Datasets involving Error Analysis}
\label{tab:mwp_datasets}
\end{table*}

\section{Related Work}
\label{sec:related_work}
\paragraph{MATH Capabilities Enhancement.}
Research on enhancing LLMs' mathematical capabilities divides into prompting-based and scaling-based approaches. Prompting techniques like Chain-of-Thought (CoT) \citep{wei2022chain}, Tree-of-Thought (ToT) \citep{yao2023tree}, and Graph-of-Thought (GoT) \citep{besta2024graph} structure reasoning paths, while Program-of-Thought (PoT) \citep{chen2023program} and Program-Aided Language Models (PAL) \citep{gao2023pal} transform reasoning into executable code. Self-Consistency \citep{wang2023selfconsistency} improves performance through sampling and probability-based selection. Scaling-based approaches focus on larger models and extensive training data \citep{azerbayev2023llemma}, exemplified by Minerva's 540B parameter model \citep{lewkowycz2022solving}. OpenAI's o1 model \citep{openai2024o1} extends scaling to test-time with external supervision \citep{setlur2024rewarding,mathshepherd}, demonstrating how scaling principles enhance mathematical reasoning during both training and inference \citep{snell2408scaling,wu2024inference}.

\paragraph{Math Word Problems.} The development of math word problem datasets has played a pivotal role in assessing the mathematical capabilities of LLMs~\citep{mathsurvey}. The evolution of these datasets reflects the progressive advancement in problem complexity and diversity. Addsub~\citep{addsub} establishes a foundation with a collection of elementary addition and subtraction problems, which is subsequently expanded by \citet{multiarith} to encompass multiplication and division operations. \citet{asdiv} further enriches the landscape by introducing a broader spectrum of text patterns and problem types, while SVAMP~\citep{svamp} innovatively applies systematic variations to existing problems to test model robustness. As model capabilities continue to advance, researchers respond by developing increasingly sophisticated datasets~\citep{LLMMathSurvey}. GSM8K~\citep{gsm8k} marks a significant step forward with its introduction of 8,500 linguistically diverse grade school math word problems. MATH~\citep{math} substantially elevates the complexity threshold by presenting 12,500 competition-level mathematics problems that span multiple domains, including algebra, probability, calculus, and geometry. TAL-SCQ5K~\citep{TAL-SCQ5K} contributes to this progression by constructing multiple-choice questions at both junior and senior high school levels. For advanced mathematical reasoning assessment, AQuA~\citep{aqua} and MathQA~\citep{mathqa} extend the frontier to college-level mathematics by incorporating GRE questions into their collections.

\paragraph{MATH Error Analysis.} Recent research has increasingly focused on analyzing and classifying errors in math word problems. Several datasets have emerged to facilitate this analysis, each employing distinct approaches to error generation and classification. MathCritique-76k~\citep{xi2024enhancingllmreasoningcritique} and EIC-MATH~\citep{EIC} utilize LLM-based approaches to generate erroneous solutions, with the latter specifically employing GPT-4 to transform correct solutions into incorrect ones. REASONEVAL~\citep{xia2024evaluating} takes a more structured approach, introducing errors through six specific perturbation strategies, including step repetition, removal, swapping, and random modifications. PRM800K~\citep{lightman2024lets} provides a comprehensive process supervision dataset with 800,000 step-level correctness labels for model-generated solutions. MR-GSM8K~\citep{MRGSM8K} and MathCheck-GSM~\citep{MATHCHECK} focus on fine-grained step-by-step error annotation, with the former incorporating specialized data augmentation strategies (PoT and REVERSED) validated through human verification.

However, these existing approaches face several critical deficiencies. First, the relatively small scale of most datasets and their limited number of error generators restrict their generalizability. Second, the synthetic nature of error construction, often involving manual injection of errors into correct solutions, may not accurately reflect the natural failure modes of LLMs. Third, current error classification schemes typically lack granularity and precise criteria, leading to ambiguous categorizations that may not effectively capture the nuanced ways in which LLMs fail at mathematical tasks. To address these limitations, we introduce MWPES-300k. As shown in Table~\ref{tab:mwp_datasets}, compared to other error analysis datasets, MWPES-300k features the most comprehensive data scale, encompasses a broader range of models and datasets, samples from naturally occurring model errors, and employs dynamic error typing that does not rely on manual categorization.

\begin{figure}[t]
  \centering
  \includegraphics[width=0.5\textwidth]{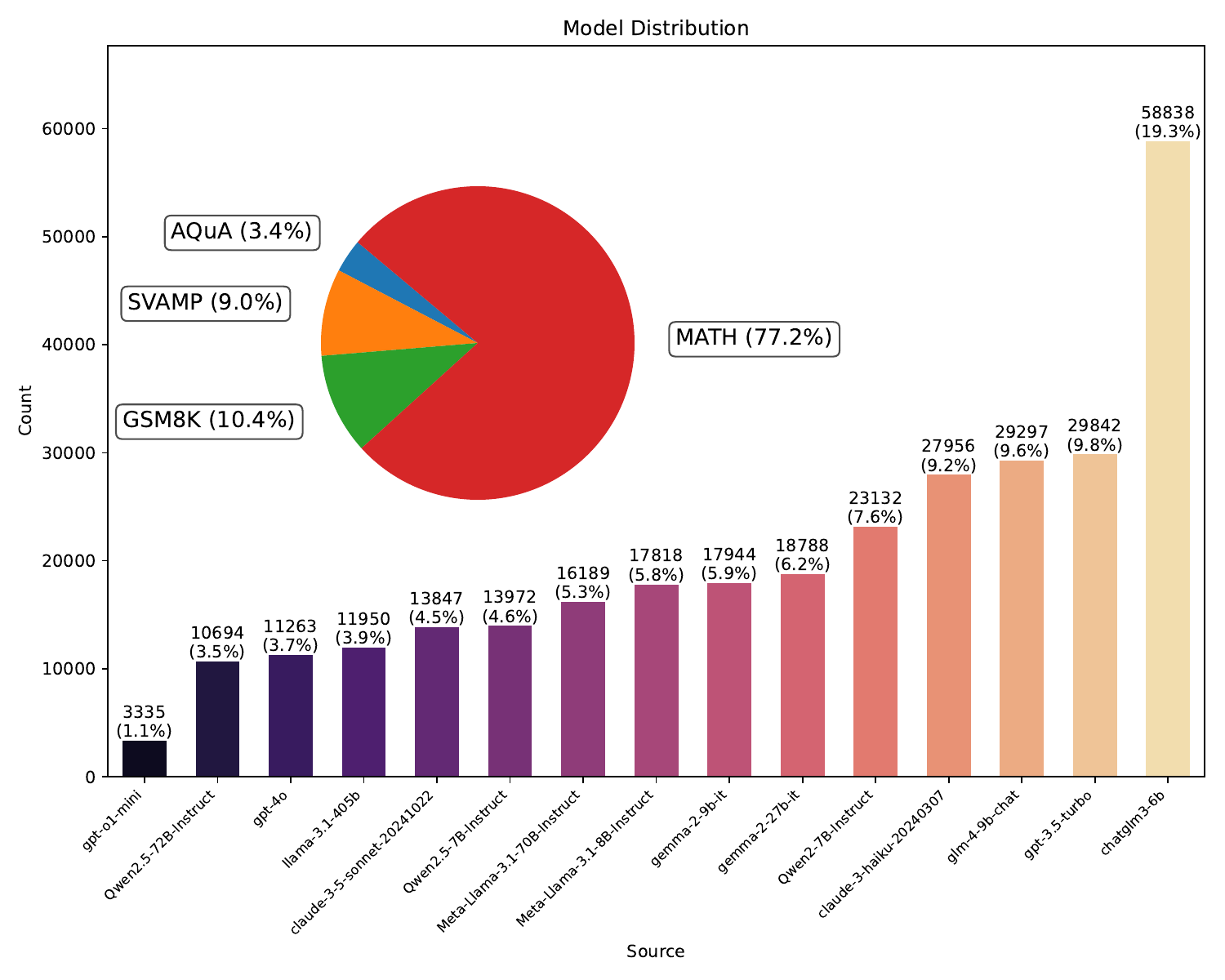}
  \caption{Distribution of sample source datasets and model sources. The pie chart illustrates the relative proportion of samples from each dataset, while the bar chart depicts the sample distribution across source models.}
  \label{fig:data-stat}
  \vspace{-1em}
\end{figure}

\section{Data Construction}
\label{sec:data_construction}
To address these deficiencies, we introduce MWPES-300K (Math Word Problem Error Solutions), a systematically constructed dataset containing 304,865 erroneous MWP solutions. Figure~\ref{fig:data-stat} presents a detailed statistical analysis of the dataset's composition and distribution. By harnessing real model outputs rather than manually injecting synthetic errors, MWPES-300K more accurately mirrors the organic failure modes of LLMs and offers a broader coverage of error types. While existing error analysis datasets like PRM800K and MR-GSM8K focus on single datasets, MWPES-300K spans four diverse MWP datasets, enabling more comprehensive analysis of error patterns across different problem types and difficulty levels. The overview of the dataset construction process is shown in Appendix~\ref{appendixDataConstruction}.

\vspace{-.5em}
\paragraph{Generation Method.}
To efficiently construct a large-scale error sample dataset, we leverage a key property of MWPs: a solution deviating from the standard answer often indicates flaws in the reasoning process~\citep{lightman2024lets}. Therefore, in generating error data, we focus on the incorrect outcomes of model solutions rather than meticulously labeling errors at each step. By avoiding manual annotation of each reasoning step, we enhance the efficiency of data generation.

We select four widely used MWP datasets: SVAMP~\citep{svamp}, GSM8K~\citep{gsm8k}, MATH~\citep{math} and AQuA~\citep{aqua}, which cover elementary, middle school, high school, and college levels questions~\citep{ahn2024large} as the data sources for MWPES-300K. We employ 15 LLMs as solution generators. These LLMs possess notable natural language understanding and generation capabilities, enabling them to simulate potential errors in the problem-solving process. For each problem, we implement a multiple sampling approach where each LLM generates 10 distinct solutions. By comparing these solutions against ground truth answers, we identify erroneous solutions that capture various mistake types, including logical errors, computational mistakes, and methodological flaws.

\vspace{-.5em}
\paragraph{Data Filtering.}
After the initial generation of the error sample set, we observe instances of misidentification. Some solutions with formatting deviations have correct reasoning and calculations. For example, an LLM might generate solutions like "\textit{Since he can only plant 6 and 3 rows respectively, then 6 * 8 blue tulips and need 3 * 6 red tulips.}" While the LLM accurately represents the computational steps, it fails to simplify "\textit{3 * 6}" to "\textit{18}," leading to its misclassification as an incorrect solution in the initial screening. These format-related deviations do not represent genuine errors and therefore require removal from the error dataset.

To address this issue, we employ an automated checker, as detailed in Appendix~\ref{appendixAutomatedChecker}, to screen all solutions initially flagged as incorrect. This checker analyzes LLM outputs to determine whether the reasoning and calculations are correct, even if the format deviates slightly from the standard answer. If the checker confirms the solution's logical validity, the corresponding sample is removed from the error sample candidate set.

Given the scale of MWPES-300K, exhaustive manual verification is impractical. To validate our automated screening method, we conduct 10 random samplings, each examining 3\% of the dataset through rigorous manual review. The average misclassification rate across these samplings remains consistently below 1\%, validating the reliability of our automated approach. After applying all filtering stages described above, we obtain our final dataset MWPES-300K.

\begin{figure*}[t]
  \centering
  \includegraphics[width=0.8\textwidth]{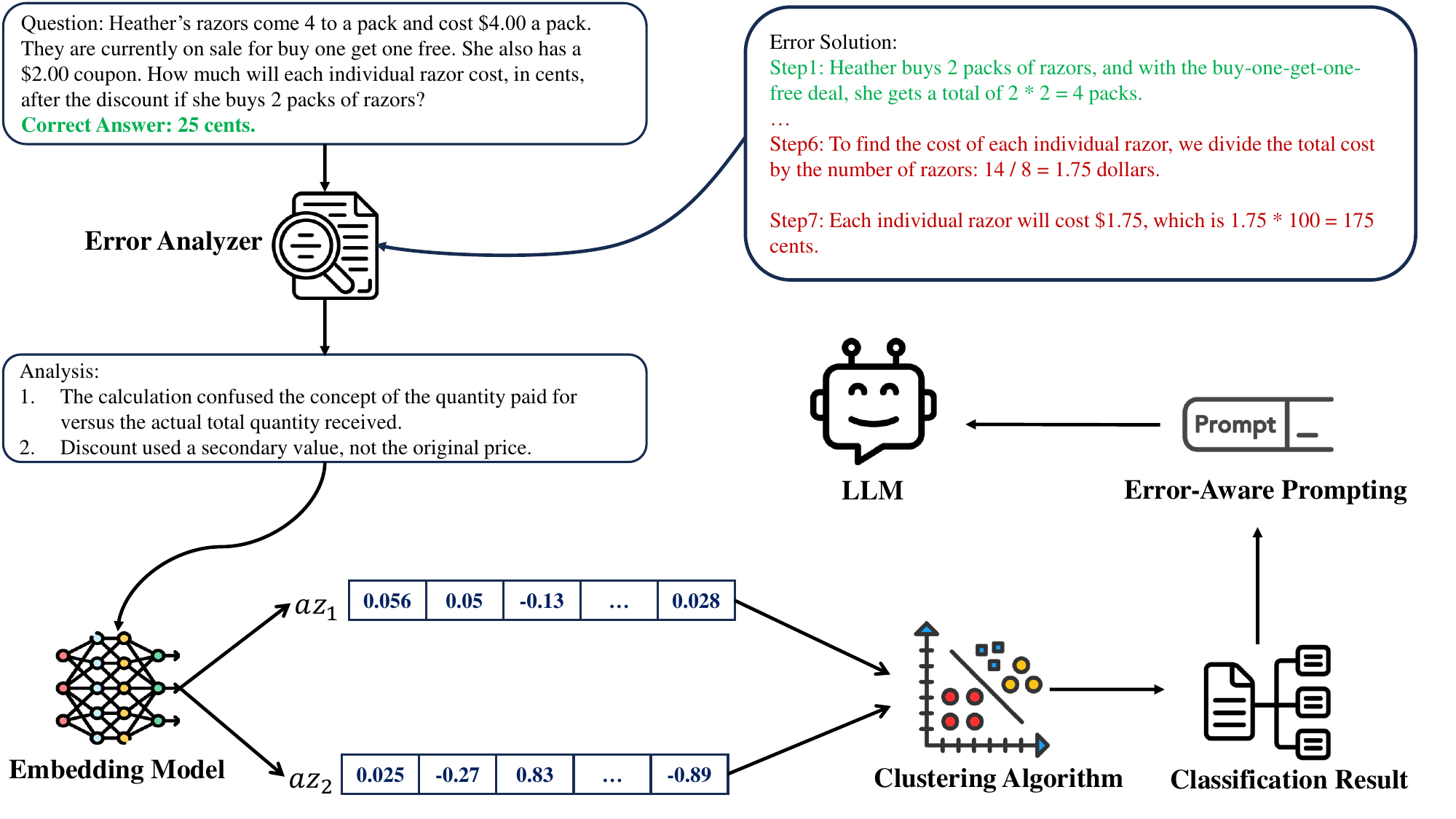}
  \caption{An illustrative example of the automated framework for dynamic error classification.}
  \label{fig:framework}
  \vspace{-1em}
\end{figure*}  

\section{Error Classification Framework}
\label{sec:Methodology}
To address the limitations of static error categorization, we develop the first automated framework for adaptive mathematical error classification. Our framework dynamically adapts to diverse error patterns while effectively maintaining classification consistency across different LLMs and problem types. The architecture of this framework is illustrated in Figure~\ref{fig:framework}.

\vspace{-.5em}
\paragraph{Data Source:} The dataset consists of a collection \( \mathcal{D} = \{(q_i, a_i, s_i)\}_{i=1}^n \), where for each index \( i \): \( q_i \) denotes the \( i \)-th question from the problem set, \( a_i \) represents its corresponding correct answer, and \( s_i \) denotes the erroneous solution.

\vspace{-.5em}
\paragraph{Error Analyzer:} The error analyzer takes a triplet \( (q_i, a_i, s_i) \) as input and outputs \( z_i \), where \( z_i = [e_{1}, e_{2}, ..., e_{m}] \) is a list structure containing \( m \) error analyses of error solution \( s_i \), when providing question \( q_i \) and standard answer \( a_i \).

\vspace{-.5em}
\paragraph{Embedding Model:} The embedding model takes the list \( z_i = [e_{1}, e_{2}, ..., e_{m}] \) as input, and for each error analyses \(e_{i}\), it outputs a corresponding vector \(v_i \in R^d\), where \(d\) is the dimension of each vector.

\vspace{-.5em}
\paragraph{Clustering Algorithm:} The clustering algorithm takes a set of vectors \( V = \{v_1, v_2, ..., v_n\} \) as input and outputs a set of \( k \) clusters, denoted as \(C = \{c_1, c_2, ..., c_k\}\). Each cluster \(c_i\) contains a set of similar error analysis vectors.

By dynamically mapping model errors into different clusters, our framework achieves adaptive error classification. Since models often exhibit multiple error types simultaneously, solutions may be mapped to several error clusters. Through comprehensive analysis of each cluster, we can derive the corresponding error typology, enabling fine-grained understanding of model failure modes.

\begin{algorithm}[t]
    \caption{Error-Aware Prompting}
    \label{alg:detailed_error_aware_prompting}
    \begin{algorithmic}[1]
        \STATE \textbf{Input:} Question $q$, Cluster centroids $C$, error analysis $z$, similarity threshold $\theta$, similarity function $\mathcal{F}$.
        \STATE Use LLM to generate knowledge point labels $K$ for $q$.
        \STATE Initialize relevant error summary set $\mathcal{R} = \emptyset$.
        
        \FORALL{$k \in K$}
            \STATE Embed $k$ into vector $v_k$.
            \FORALL{$c \in C$}
                \FORALL{$v_e \in c.\textit{summaries}$}
                    \STATE Calculate $\mathcal{F}(v_k, v_e)$.
                    \IF{$\mathcal{F}(v_k, v_e) \geq \theta$}
                        \STATE Add $e$ to $\mathcal{R}$.
                    \ENDIF
                \ENDFOR
            \ENDFOR
        \ENDFOR
        \STATE Construct prompt $p$ by appending $R$ to $q$.
        \STATE Generate response using $p$ with LLM.
        \STATE \textbf{Output:} LLM response.
    \end{algorithmic}
\end{algorithm}

\section{Error-Aware Prompting}
\label{sec:eap}
The Error-Aware Prompting algorithm aims to enhance the LLM's ability by incorporating relevant errors of the question. This algorithm involves two main steps: Knowledge Point Labeling and Relevant Error Summary Retrieval.

\begin{itemize}
    \vspace{-.5em}

    \item \textbf{Knowledge Point Labeling:} Given a question \( q \), an LLM is employed to generate a set of knowledge point labels \( K \). These labels represent the relevant concepts or topics involved in the question.
    \vspace{-.5em}

    \item \textbf{Relevant Error Summary Retrieval:} For each knowledge point label \( k \in K\), we first embed \(k\) into a vector \(v_k\).  We then iterate through the clustering results \(C\), where each cluster \( c_i \in C\) contains embedded error summaries. A similarity function \(\mathcal{F}(v_k, v_e)\) is used to determine the relevance between the knowledge point \(k\) and the error summary \(e\). If \(\mathcal{F}(v_k, v_s) \geq \theta\), where \(\theta\) is a predefined similarity threshold, the error summary \(s\) is added to a set of relevant summaries \(R\).
    \vspace{-.5em}
    
\end{itemize}

An error-aware prompt \(p\) is constructed by appending the relevant summaries \(R\) to the original question \(q\). This prompt is then fed into the LLM to generate a response. The detail of the algorithm is shown in Algorithm~\ref{alg:detailed_error_aware_prompting}.

\section{Experiment}
\label{sec:experiment}

\begin{figure*}[h]
  \centering
  \includegraphics[width=0.8\textwidth]{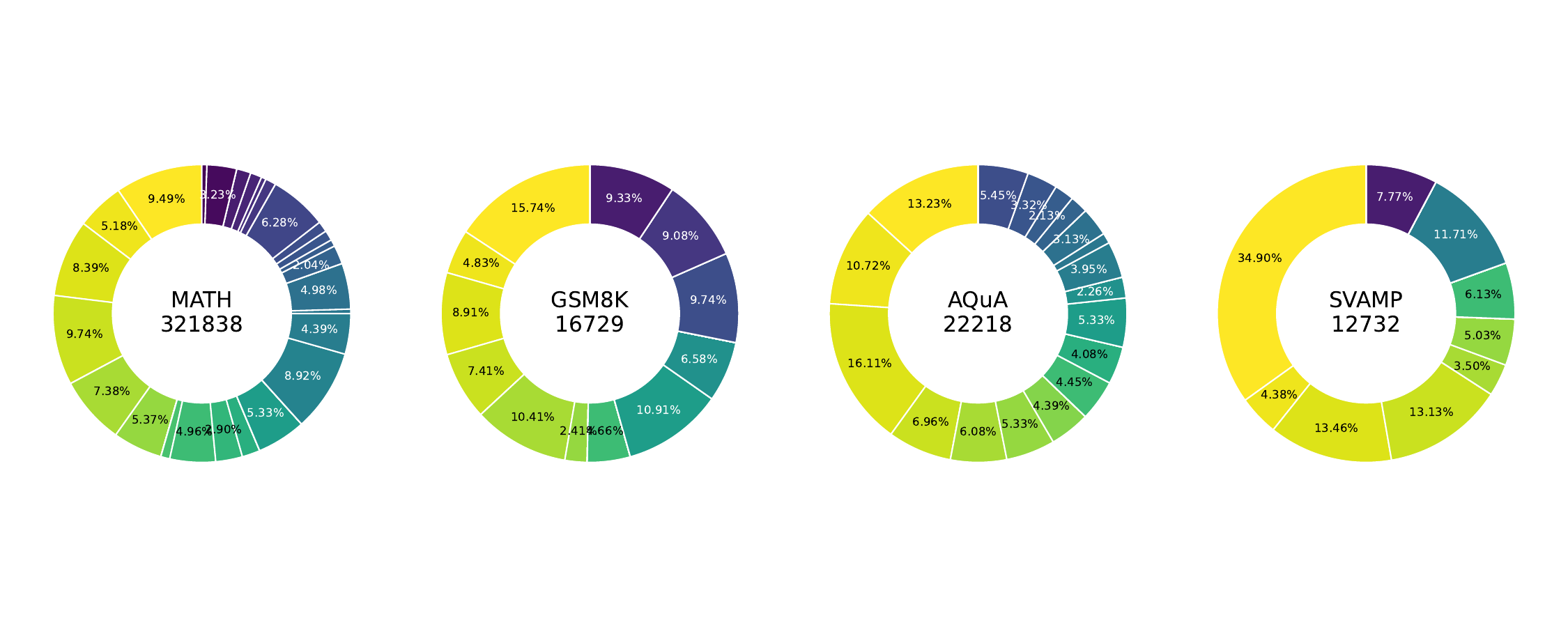}
  \vspace{-3em}
  \caption{The error pattern distributions of GPT-4o across 4 MWP datasets. Each solution may make more than one errors.}
  \label{fig:rq1}
  \vspace{-1em}
\end{figure*}

\begin{figure*}[h]
  \centering
  \includegraphics[width=0.8\textwidth]{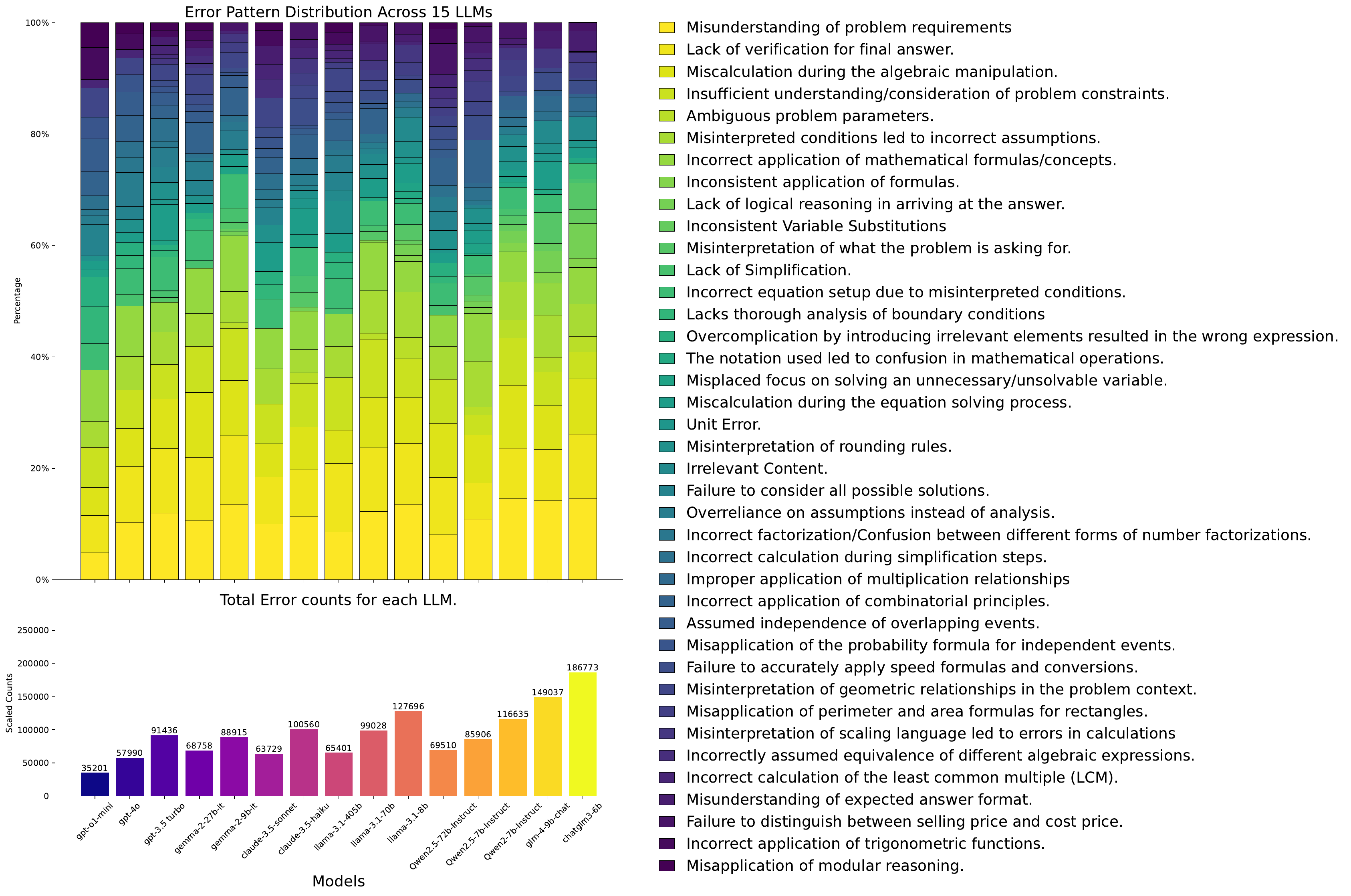}
  \vspace{-1em}
  \caption{The figure depicts error pattern distributions for 15 LLMs. The top stacked bar chart shows the distribution of error types per model, while the bottom bar chart shows the total error counts. A color-coded legend to the right shows error types descriptions.}
  \label{fig:rq2-3}
  \vspace{-1em}
\end{figure*}

\subsection{Experiment Setup}

\paragraph{Data Source:}
We utilize the novel MWPES-300K dataset developed in this research.

\paragraph{Error Analyzer:}
We employ GPT-4o to analyze the causes of errors in each solution path. The input is the error sample, and the output is a description and analysis of the error type. The prompt template used is detailed in the Appendix Table~\ref{tab:prompt4gptErrorAnalyzer}. We also investigate the framework's adaptability to open-source models, as demonstrated in Appendix~\ref{appendix:alternative_analyzers}.

\paragraph{Text Embedding:}
To convert textual error descriptions into numerical vectors, we use OpenAI's \texttt{text-embedding-3-large} ~\citep{openai2024embedding}. 

\paragraph{Clustering Algorithm:}
We apply K-means as the clustering algorithm. We conduct an ablation study for the selection (See Appendix~\ref{sec:appendix_clustering_ablation}). To determine the optimal number of clusters \textit{k}, we adopt 3 methods: the Davies-Bouldin Index~\citep{davies1979cluster}, Gap Statistic~\citep{tibshirani2001estimating}, and the Elbow Method. Through this multi-stage validation process, we identify $k=39$ as the optimal number of error categories. The detailed selection procedure is presented in Appendix~\ref{appendix:k_selection}.

\subsection{Main Results}
We conduct extensive experiments to address the following research questions:

\paragraph{RQ1: How do different MWP datasets impact error patterns for a given model?} As presented in Figure~\ref{fig:rq1}, our analysis of error patterns across GPT-4o on MATH, GSM8K, AQuA, and SVAMP datasets reveals two key findings. First, dataset difficulty strongly correlates with error type diversity, with the more challenging MATH dataset consistently eliciting both higher error counts and more diverse error types compared to the simpler SVAMP and GSM8K datasets. Second, certain common error patterns transcend specific problem types across all datasets, including \textit{"misinterpretation of problem requirements"}, \textit{"algebraic manipulation errors"}, and \textit{"incomplete constraint consideration"}. Notably, errors in the SVAMP dataset frequently stem from ambiguous problem descriptions rather than model weaknesses.

\paragraph{RQ2: How do error patterns change as model capabilities improve?} In Figure~\ref{fig:rq2-3}, analysis across model generations reveals a significant evolution in mathematical reasoning capabilities. Within the GPT series, we observe a consistent reduction in fundamental mathematical errors, with \textit{"Misunderstanding of problem requirements"} decreasing from 11.93\% (10,907 errors) in GPT-3.5-turbo to 4.83\% (1,699 errors) in o1-mini. Similar improvements appear in algebraic manipulation errors. Interestingly, as models advance, their error patterns shift from basic computational mistakes to more sophisticated reasoning failures. This is exemplified by the emergence of \textit{"Lacks thorough analysis of boundary conditions"} errors, which are absent in GPT-3.5-turbo but appear at a rate of 2.33\% in GPT-4o, suggesting an evolution toward more nuanced mathematical reasoning.

\paragraph{RQ3: How do error patterns change as the parameter size of LLM increases?} As illustrated in Figure~\ref{fig:rq2-3}, horizontal comparison across Llama, Gemma-2, and Qwen2.5 families reveals a consistent trend: total error counts decrease as model size increases, indicating a direct correlation between scale and performance. For instance, within the Llama series, Llama-3.1-405b (65,401 total errors) significantly outperforms Llama-3.1-8b (127,696 errors), with \textit{"Misunderstanding of problem requirements"} errors decreasing from 13.56\% to 8.58\%. Additionally, certain error types like \textit{"Inconsistent Variable Substitutions"} become almost non-existent in larger models. Furthermore, larger models exhibit more balanced error distributions, suggesting fewer prominent weaknesses compared to their smaller counterparts.

\paragraph{RQ4: Can dynamic error classification effectively enhance model output consistency and accuracy?} In Table~\ref{tab:classification_consistency}, we analyze the accuracy and consistency of error classification using both our dynamic framework and static typologies. We measure consistency using $1-H(x)$, where $H(x)$ represents the entropy of prediction results across multiple samples. Using 10 samples to calculate entropy and average accuracy, along with a standardized set of 9 error types for fair comparison, we observe that predefined static templates yield lower accuracy and consistency due to ambiguous and overly broad error type definitions. In contrast, our adaptive error classification framework leverages the model's inherent analytical and comprehension capabilities to significantly enhance both the accuracy and consistency of error classification. Detailed analysis of this phenomenon is presented in Appendix~\ref{appendixStaticErrorCategorization}.

\begin{table}[t]
\centering
\begin{tblr}{
  column{2} = {c},
  column{3} = {c},
  hline{1,6} = {-}{0.08em},
  hline{2} = {-}{},
}
        \textbf{Model} & \textbf{Accuracy} & \textbf{Consistency} \\
        GPT-4o (Stastic) & 35.65\% & 18.27\% \\
        GPT-4o (Adaptive) & 83.33\% & 82.21\% \\
        \hline
        Claude-3.5 (Stastic) & 27.45\% & 12.63\% \\
        Claude-3.5 (Adaptive) & 74.57\% & 76.51\% \\
    \end{tblr}
    \caption{Models using the adaptive framework demonstrate significant improvements in both classification accuracy and consistency compared to static categories.}
    \label{tab:classification_consistency}
\vspace{-1em}
\end{table}

\begin{table}[t]
\centering
\begin{tblr}{
  column{2} = {c},
  column{3} = {c},
  hline{1,6} = {-}{0.08em},
  hline{2} = {-}{},
}
    \textbf{Model} & \textbf{CoT} & \textbf{EAP} \\
    GPT-4o & 76.60\% & 79.22\%$_{\color{red}{+2.62\%}}$ \\
    Claude-3.5 & 71.10\% & 80.02\%$_{\color{red}{+8.92\%}}$ \\
    Llama-3.1-70b & 68.0\% & 78.48\%$_{\color{red}{+10.48\%}}$ \\
    Llama-3.1-8b & 51.90\% & 78.23\%$_{\color{red}{+26.33\%}}$ \\
    \end{tblr}
    \caption{Performance comparison between commercial and open-source models using CoT baseline and Error-Aware Prompting on the MATH dataset.}
    \label{tab:indomain}
\vspace{-1em}
\end{table}

\subsection{Performance of EAP}
\label{subsec:error_aware_prompting}
We evaluate the effectiveness of Error-Aware Prompting (EAP) on both in-distribution and out-of-distribution MWPs to investigate whether problem-specific error feedback could enhance LLM performance. Our experiments focuses on two challenging datasets: MATH~\citep{math} and TAL-SCQ5K~\citep{TAL-SCQ5K}. For each problem, we employ the approach detailed in Algorithm~\ref{alg:detailed_error_aware_prompting} to retrieve relevant error summaries. These summaries, generated by the error analyzer, detail the mistakes observed and are appended to the original prompt to guide the model's reasoning process.

\paragraph{In-Distribution Analysis} 
We evaluate 4 LLMs on the MATH dataset. As shown in Table~\ref{tab:indomain}, Error-Aware Prompting significantly improved the accuracy across all models. For instance, Llama-3.1-8b demonstrated a performance increase of over 25\%, surpassing the CoT performance of GPT-4o. These findings highlight that EAP effectively helps LLMs avoid potential errors, leading to substantial improvements in performance. Crucially, as demonstrated in Appendix~\ref{appendix:static_baseline}, this improvement stems from our dynamic error selection mechanism rather than merely exposing models to error examples—static error baselines actually degrade performance compared to standard CoT.

\begin{figure}[t]
  \centering
  \includegraphics[width=0.5\textwidth]{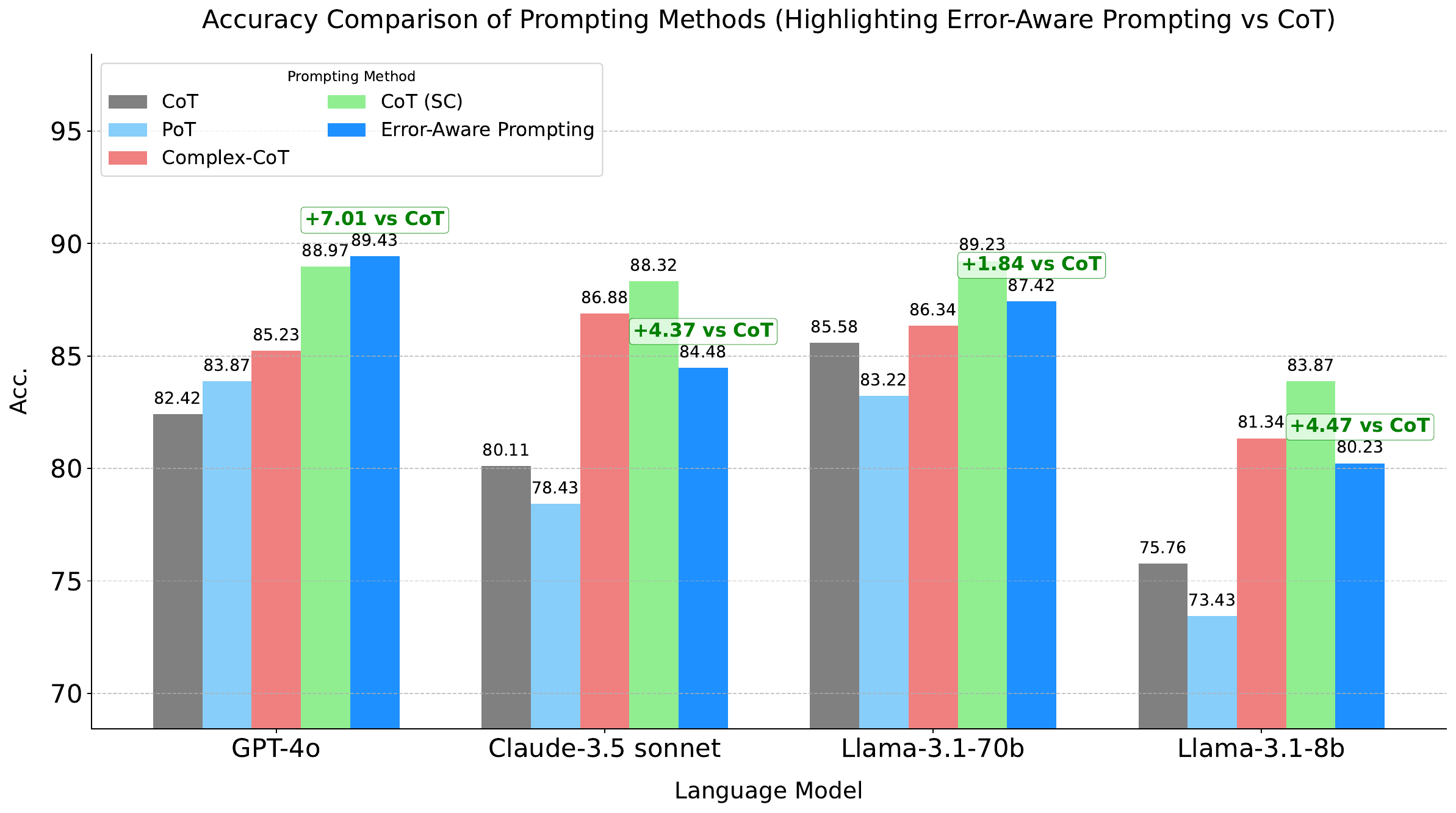}
  \caption{Performance comparison of prompting methods on 4 LLMs, highlighting Error-Aware Prompting's (EAP) improvement over CoT.}
  \label{fig:prompt-comparison-chart}
  \vspace{-1em}
\end{figure}

\paragraph{Out-of-Distribution Analysis} 
To comprehensively validate the generalization capability of Error-Aware Prompting, we conduct two types of out-of-distribution (OOD) experiments: dataset-level OOD to test cross-dataset transfer and model-level OOD to verify applicability to unseen models.

\paragraph{Dataset-Level OOD.}
We first evaluate EAP on TAL-SCQ5K~\citep{TAL-SCQ5K}, a dataset entirely disjoint from our MWPES-300K training corpus. TAL-SCQ5K consists of multiple-choice questions spanning primary to high school mathematics, providing a distinct problem format from the free-response datasets (MATH, GSM8K, SVAMP) used in MWPES-300K construction. For each problem, we extract its knowledge point route field representing the hierarchical structure of mathematical concepts. We employ TF-IDF for knowledge point embedding, followed by cosine similarity matching to retrieve relevant error summaries.

As shown in Figure~\ref{fig:prompt-comparison-chart}, EAP demonstrates substantial performance improvements over CoT across all models on TAL-SCQ5K. GPT-4o exhibits more than 7\% improvement, surpassing even CoT with self-consistency~\citep{wang2023selfconsistency}. This strong performance on an entirely unseen dataset format indicates that the error patterns captured by our framework transfer effectively across different problem types and difficulty levels.

\paragraph{Model-Level OOD.}
To verify that our error patterns generalize beyond the 15 source models used in MWPES-300K construction, we evaluate EAP on the Qwen3 model family~\citep{yang2025qwen3}. Qwen3 models contributed no error samples to our MWPES-300K dataset. As shown in Table~\ref{tab:model_ood}, EAP consistently improves performance across all three Qwen3 variants on the MATH dataset.

\begin{table}[t]
\centering
\begin{tblr}{
  column{2} = {c},
  column{3} = {c},
  column{4} = {c},
  hline{1,5} = {-}{0.08em},
  hline{2} = {-}{},
}
    \textbf{Model} & \textbf{CoT} & \textbf{EAP}\\
    Qwen3-4B & 81.8\% & 87.7\%$_{\color{red}{+5.9\%}}$ \\
    Qwen3-8B & 84.2\% & 88.1\%$_{\color{red}{+3.9\%}}$ \\
    Qwen3-14B & 87.5\% & 91.2\%$_{\color{red}{+3.7\%}}$ \\
\end{tblr}
\caption{Performance of Error-Aware Prompting on unseen Qwen3 models. These models did not contribute to MWPES-300K construction.}
\label{tab:model_ood}
\vspace{-1em}
\end{table}

The results reveal that smaller models benefit most from error guidance, with Qwen3-4B achieving a 5.9\% improvement. This suggests that models more prone to common reasoning errors gain greater value from explicit error patterns. Even the largest Qwen3-14B model shows meaningful gains (+3.7\%), indicating that our framework captures fundamental error patterns that transcend specific model architectures.

These OOD experiments, spanning both unseen datasets and unseen models—demonstrate that the error patterns identified through our dynamic classification framework represent genuine, transferable insights into mathematical reasoning failures rather than dataset-specific or model-specific artifacts. Further detailed analyses are presented in Appendix~\ref{appendixDetailOODAnalysis}, and we analyze EAP's impact on specific error categories in Appendix~\ref{sec:per-error-type-analysis}.

\paragraph{Impact of EAP on Reasoning Strategies}
To investigate how EAP influences LLMs' reasoning strategies, we conducted a qualitative analysis using 100 randomly selected problems from the MATH dataset. We compared the problem-solving approaches of GPT-4o and Llama-3.1-70b under CoT prompting versus our EAP method.

\begin{table}[t]
\centering
\begin{tabular}{@{}lcc@{}}
\toprule
Model           & Self-Reflection Rate \\ \midrule
GPT-4o (CoT)              & 0.0\%                         \\
GPT-4o (EAP)              & 73.0\%                        \\ \midrule
Llama-3.1-70b (CoT)       & 2.0\%                         \\
Llama-3.1-70b (EAP)       & 54.0\%                        \\ \bottomrule
\end{tabular}
\caption{Observed Self-Reflection Rates in Reasoning Steps for CoT vs. Error-Aware Prompting (EAP) on a subset of 100 MATH dataset samples.}
\label{tab:self_reflection_rates}
\vspace{-1em}
\end{table}

As presented in Table~\ref{tab:self_reflection_rates}, our analysis revealed that EAP effectively induces self-reflection~\citep{kumar2025training}, enabling models to exhibit self-assessment and self-correction behaviors during solution generation. For GPT-4o, compared to CoT, the likelihood of model self-reflection increased by 73.0\%. In the open-source Llama-3.1-70b model, the rate of reflection rose dramatically from 2.0\% to 54.0\%. Further observations indicated that through EAP, models become cognizant of potential biases or errors in their initial or intermediate reasoning steps (as indicated by the error summaries) and subsequently attempt to rectify these issues in the ensuing solution steps. This aligns with our earlier findings regarding error pattern evolution across model capabilities, where more advanced models demonstrate greater capacity for nuanced reasoning. Interestingly, when models are highly confident in their outputs, introducing EAP does not lead to significant divergence from CoT baseline reasoning paths. This selective intervention suggests that EAP acts as an adaptive mechanism, providing guidance primarily when needed.

\section{Conclusion}

We introduce a dynamically adaptive framework for error classification in MWP solving. It automatically updates error categories based on observed outputs, rather than using static categories. We propose the MWPES-300K dataset, with 304,865 error samples across 15 LLMs and 4 MWP datasets, enabling a thorough error analysis. Specifically, dataset complexity positively correlates with error diversity. As LLMs enhance their core mathematical skills, the errors they make tend to become more complex, reflecting challenges in advanced reasoning. Increased model parameter size correlates with reduced overall errors and more balanced distributions. We propose Error-Aware Prompting (EAP), which leverages categorized errors to significantly improve both in-distribution and out-of-distribution performance on MATH and TAL-SCQ5K datasets. We suggest that a focus on error analysis can serve as a good way to develop more robust and reliable LLMs for mathematical reasoning tasks.

\section*{Limitations}
\label{sec:conclusion}
\paragraph{Process-Answer Inconsistency}
Our analysis focuses solely on solutions where both the final answer and solution process are incorrect, which ignores the phenomenon where LLMs arrive at correct answers through flawed reasoning processes. We leave the investigation of this type of error for future research.

\paragraph{Domain Specificity}
Although we demonstrate the effectiveness of our framework in the mathematical domain, its applicability to other domains remains unexplored. Our dynamic error classification approach could be adapted for tasks such as logical reasoning, scientific problem-solving, or code debugging. Future work will extend our framework and error-aware prompting algorithm to additional domains to analyze LLMs' error patterns across different fields more comprehensively.

\section*{Ethics Statement}
The MWPES-300K dataset comprises error samples collected from 15 language models across four established mathematical word problem datasets. We strictly adhere to all dataset licenses (Table~\ref{tab:dataset_statistic}) and data collection protocols. To ensure quality, three mathematical experts with advanced degrees reviewed the automated classifications at standard academic rates. We prioritized gender equality and regional diversity in recruitment, with detailed annotation instructions provided in Table~\ref{tab:instruction_separated}.

 All experiments with commercial LLMs are conducted in full compliance with their respective providers' terms of service and API usage policies. For open-source models, we carefully follow the associated model licenses and usage restrictions. The four source MWP datasets are utilized in accordance with their original licenses, and all derived error samples maintain proper attribution to their source problems. Our automated error classification framework is designed with careful consideration of potential biases. We implement rigorous testing procedures to ensure that the dynamic categorization system does not inadvertently perpetuate or amplify existing biases in model outputs. During our development process, we utilized GPT-4o for automated data annotation and GitHub Copilot for code writing assistance.

The MWPES-300K dataset will be released under CC-BY-SA-4.0 for research purposes only. While our work aims to improve mathematical reasoning in LLMs, we acknowledge the responsibility to prevent harmful applications and encourage researchers to maintain high ethical standards when utilizing our framework.

\subsection*{Acknowledgements}

This work is supported by National Natural Science Foundation of China NSFC (No.92370201, No. 62272165) and Shanghai Trusted Industry Internet Software Collaborative Innovation Center.

\bibliography{custom}

\clearpage

\appendix

\section{Appendix}
\label{sec:appendix}

\begin{figure*}[h]
  \centering
  \includegraphics[width=1.0\textwidth]{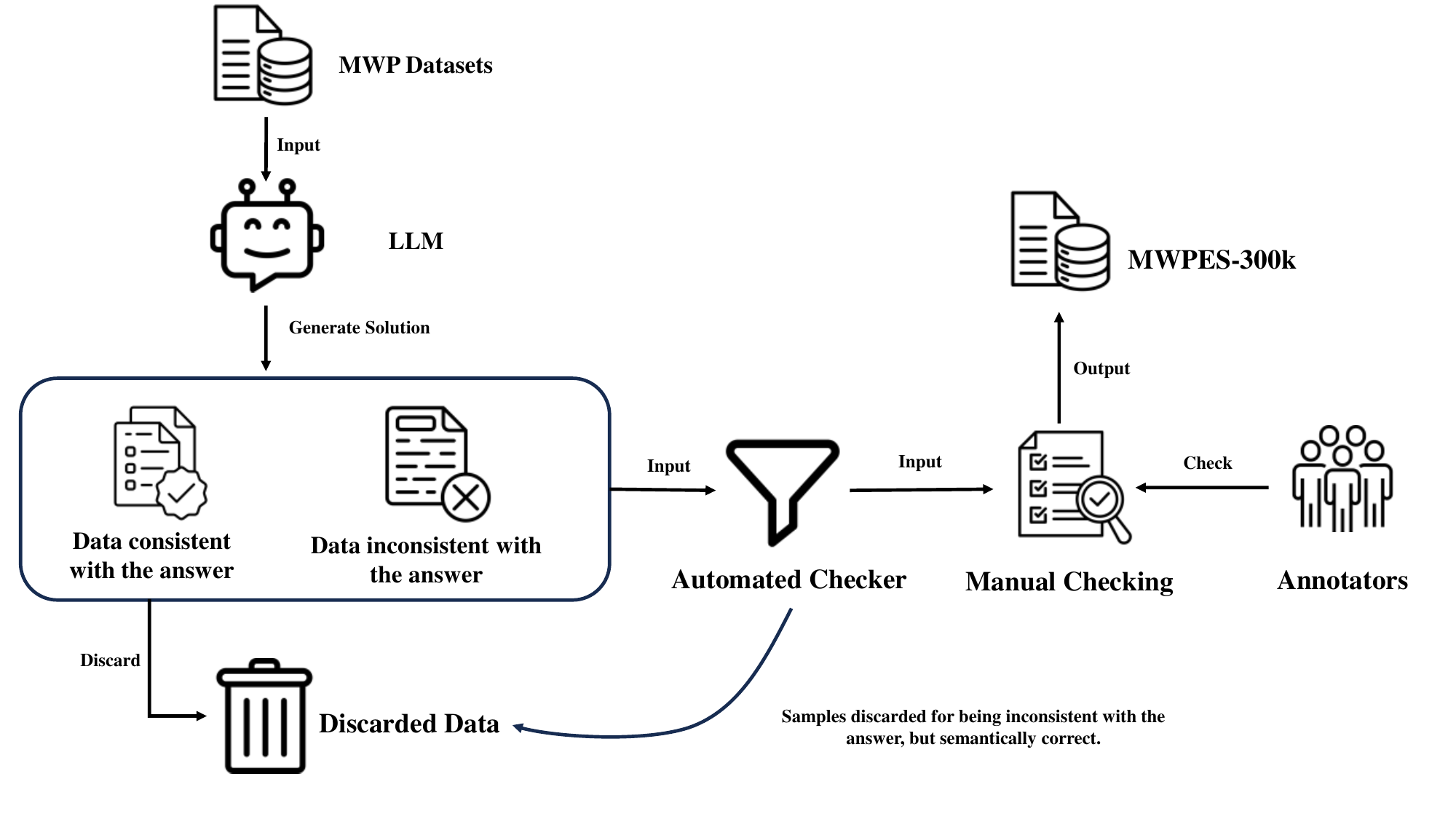}
  \caption{The overview of the dataset construction workflow.}
  \label{fig:data_construction_detail}
  \vspace{-0.5em}
\end{figure*}

\begin{table*}[t]
\centering
\footnotesize
\begin{tabular}{l|c|c|c}
\toprule
\textsc{Dataset} & \textsc{Answer Format} &  \textsc{Number} & \textsc{License} \\
\midrule
\href{https://github.com/openai/grade-school-math}{GSM8K}~\citep{gsm8k}& Number & 1,319 & MIT License \\
\href{https://github.com/wangxr14/Algebraic-Word-Problem-Solver}{MultiArith}~\citep{multiarith}& Number &  600 & Unspecified \\
\href{https://github.com/arkilpatel/SVAMP}{SVAMP}~\citep{svamp}& Number & 1,000 & MIT License \\
\href{https://github.com/hendrycks/math/}{MATH}~\citep{math}& Number & 5,000 & MIT license \\
\href{https://github.com/deepmind/AQuA}{AQUA}~\citep{aqua}& Multi-choice & 254 & Apache-2.0 \\
\href{https://github.com/math-eval/TAL-SCQ5K}{TAL-SCQ5K}~\citep{TAL-SCQ5K}& Multi-choice & 2,000 & MIT license \\
\bottomrule
\end{tabular}
\caption{Statistical overview of datasets used in data construction and experimental analysis.}
\label{tab:dataset_statistic}
\vspace{-1em}
\end{table*}
\subsection{Details of data construction}
\label{appendixDataConstruction}

Figure~\ref{fig:data_construction_detail} illustrates the data construction process. We curate a dataset by selecting 4 Math Word Problem (MWP) datasets (SVAMP, GSM8K, AQuA, and MATH) as our source. Table~\ref{tab:dataset_statistic} provides comprehensive statistics for each dataset along with their respective licenses. Throughout the process, we strictly adhered to these licenses, ensuring use consistent with their intended purposes. 15 LLMs from various vendors generated 10 solutions for each problem. These solutions are filtered based on consistency with expected answers, with inconsistent yet semantically correct data being discarded. We adopt GPT-4 as the automated checker. The remaining data are manually reviewed to create the final MWPES-300k dataset. The MWPES-300k dataset consists of English math word problems with corresponding solutions and analyses. Following our rigorous review process, we ensure the dataset is free of any personally identifying information or offensive content.

We obtain the LLM's outputs for MWP through API calls, using the default settings for each model. Specifically, the \texttt{max\_tokens} parameter, limiting the maximum length of the generated response (including the prompt), is set to \texttt{2048}. The \texttt{temperature} parameter, controlling the randomness and creativity of the generated text, is set to \texttt{0.7}.  The \texttt{top\_p} parameter is set to \texttt{1.0}. Both \texttt{frequency\_penalty} and \texttt{presence\_penalty}, designed to penalize token repetition and the introduction of new topics respectively, are set to \texttt{0.0}.

\subsection{Automated Checker}
\label{appendixAutomatedChecker}
Within the data filtering process, an automated checker plays a crucial role in refining the initially identified error sample set. This checker can take the form of any model designed to assess the validity of a MWP solving process. However, for simplicity and efficiency in this study, we utilize GPT-4o as our automated checker. The checker is prompted to evaluate only the final step of the solution, determining if it aligns semantically with the standard answer. Critically, the checker is instructed to disregard minor format or presentation differences, recognizing as correct those solutions that achieve the proper meaning despite such variations. This approach enables us to effectively filter out samples flagged due to superficial deviation.

\subsection{Details of k Selection Process}
\label{appendix:k_selection}

\paragraph{Elbow Method for Initial Range Identification}
To determine the optimal number of clusters $k$ for our error categorization framework, we employ a three-stage validation process. The first stage utilizes the Elbow Method to rapidly narrow down the search space. We compute the Within-Cluster Sum of Squares (WCSS) for $k$ values ranging from 1 to 50:
\begin{equation}
    WCSS(k) = \sum_{i=1}^{k} \sum_{x \in C_i} ||x - \mu_i||^2
\end{equation}
where $C_i$ represents the $i$-th cluster, $\mu_i$ is the centroid of cluster $C_i$, and $x$ represents data points within the cluster. By plotting the $k$-WCSS curve and identifying the elbow point where the rate of WCSS decrease significantly diminishes, we identify a candidate range of $k \in [35, 45]$. This initial screening efficiently reduces the computational burden for subsequent validation stages.

\paragraph{Gap Statistic for Statistical Validation}
Within the candidate range identified by the Elbow Method, we apply the Gap Statistic to provide rigorous statistical validation. This method compares the observed WCSS with the expected WCSS under a null reference distribution:
\begin{equation}
    Gap(k) = E_n^*[\log(WCSS_k)] - \log(WCSS_k)
\end{equation}
where $E_n^*[\log(WCSS_k)]$ is the expected value under the null hypothesis of uniform reference distribution. To assess the stability of our selection, we compute the standard error $SE_k$ across $B=100$ reference datasets using the sample standard deviation of $\log(WCSS_k^*)$ multiplied by $\sqrt{1 + 1/B}$. We apply the one standard error rule, selecting the smallest $k$ such that $Gap(k) \geq Gap(k+1) - SE_{k+1}$. This analysis indicates $k=38$ as the statistically optimal choice within our candidate range.

\paragraph{Davies-Bouldin Index for Final Optimization}
For final refinement, we evaluate clustering quality using the Davies-Bouldin Index (DBI) in the neighborhood of the Gap Statistic suggestion. The DBI measures the average similarity ratio between clusters:
\begin{equation}
    DBI = \frac{1}{k}\sum_{i=1}^{k}\max_{j \neq i}\left(\frac{\sigma_i + \sigma_j}{d(c_i, c_j)}\right)
\end{equation}
where $\sigma_i = \frac{1}{|C_i|}\sum_{x \in C_i}||x - c_i||$ represents the average within-cluster distance and $d(c_i, c_j) = ||c_i - c_j||$ denotes the distance between cluster centroids. Lower DBI values indicate better clustering quality with more compact and well-separated clusters. Evaluating $k$ values in the range $[36, 42]$, we find that $k=39$ yields the minimum DBI value. We verify this selection's stability through 10 independent runs with different random initializations, confirming $k=39$ as our final choice for the number of error categories in the MWPES-300K dataset analysis.

\subsection{Details of Static Error Categorization}
\label{appendixStaticErrorCategorization}

We design an experiment where an LLM performs repeated 9-class classifications on solutions containing only one error. These error types, as proposed by \citet{EIC}, are distilled from previous research, including numerical errors such as \textit{calculation errors (CA)}, \textit{counting errors (CO)}, and \textit{unit conversion errors (UC)}; contextual errors like \textit{context value errors (CV)} and \textit{hallucinations (HA)}; and reasoning errors such as \textit{operator errors (OP)}, \textit{formula confusion errors (FC)}, \textit{missing steps (MS)}, and \textit{contradictory steps (CS)}. 

\begin{figure*}[h]
  \centering
  \includegraphics[width=1.0\textwidth]{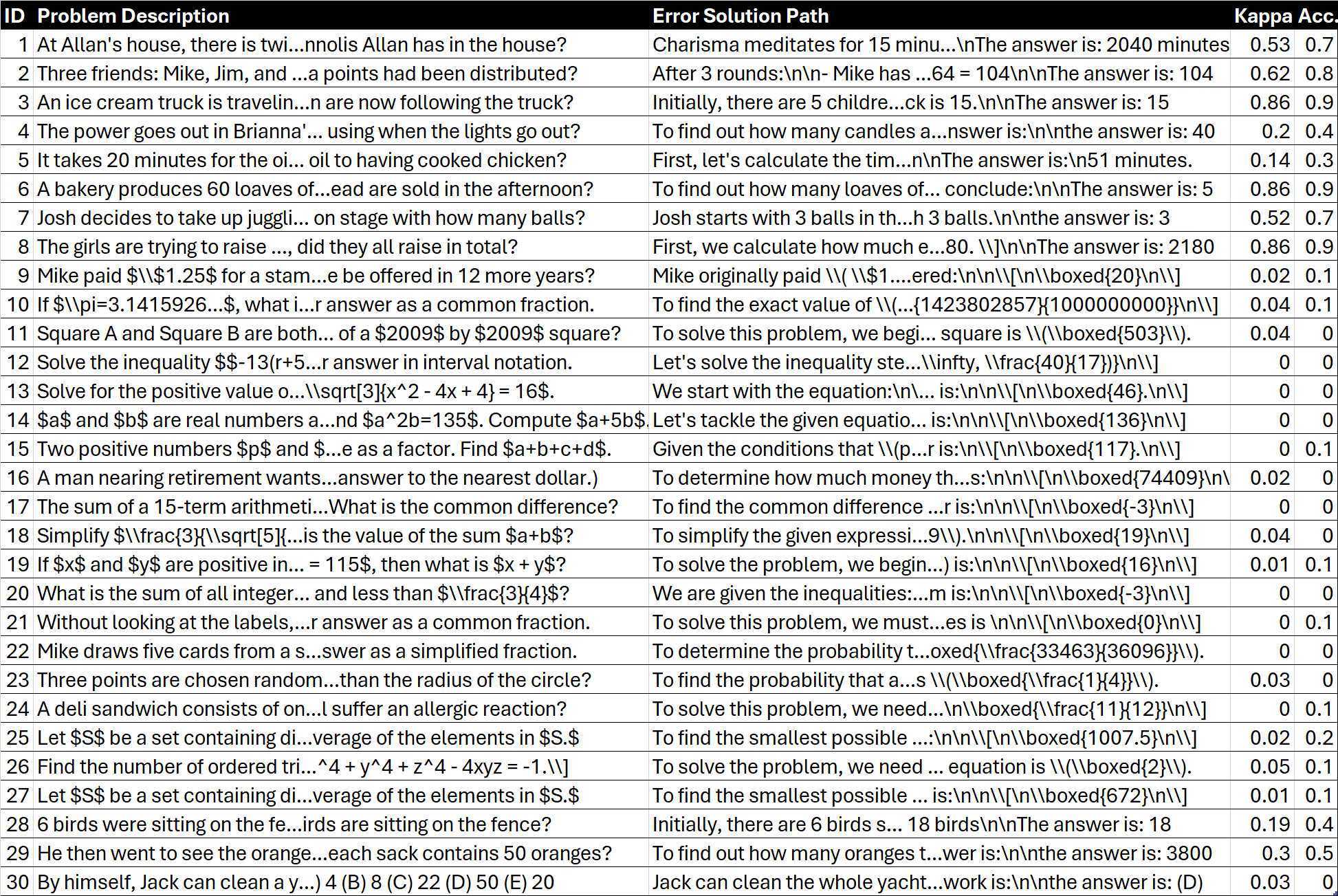}
  \caption{A breakdown example of GPT-4o's performance in classifying 9 distinct error types (CA, CO, UC, CV, HA, OP, FC, MS, CS) for 30 MWP.}
  \label{fig:GPT-4o-consistencyExperiment}
  \vspace{-0.5em}
\end{figure*}

\begin{figure*}[h]
  \centering
  \includegraphics[width=1.0\textwidth]{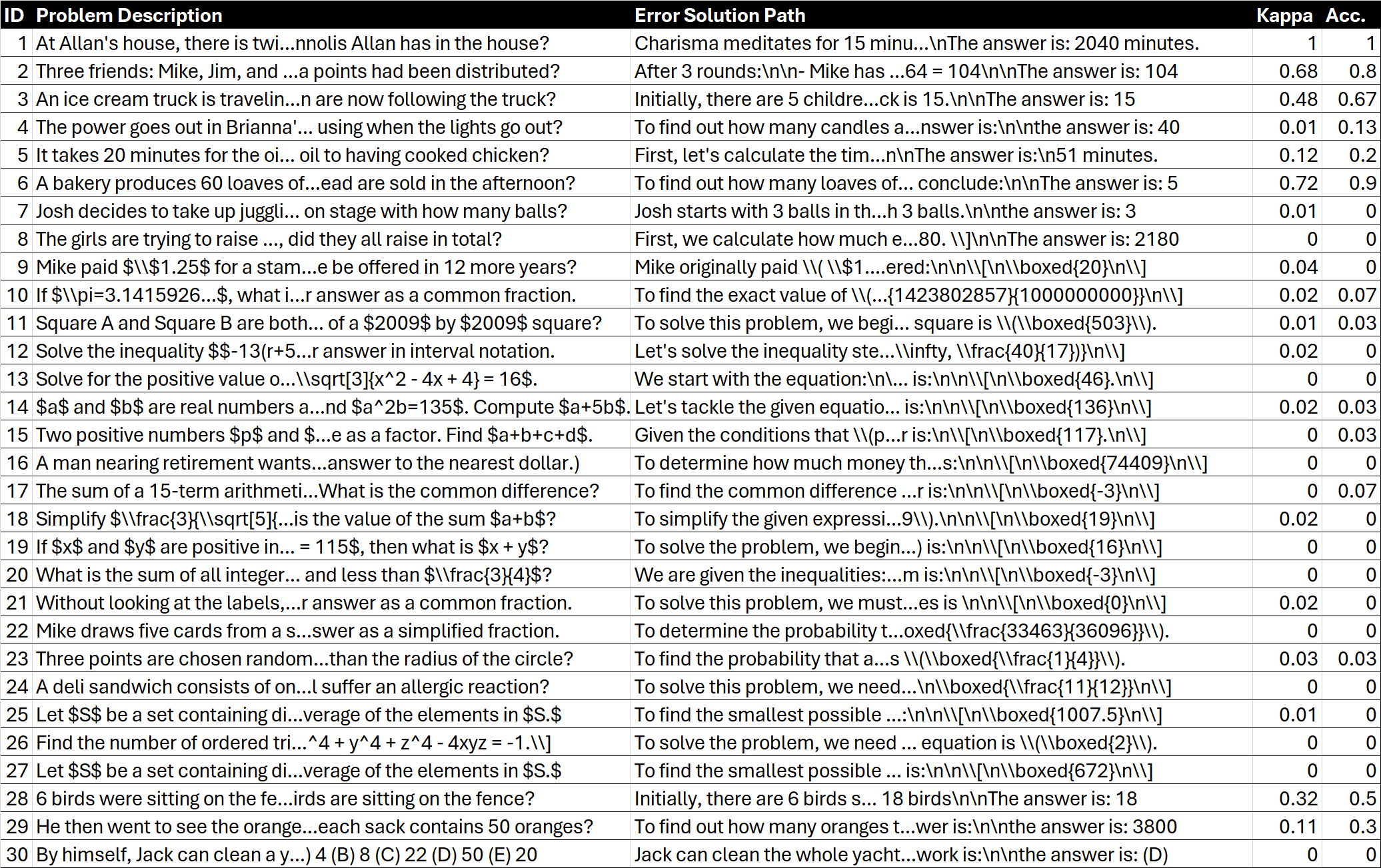}
  \caption{A breakdown example of Claude-3.5-sonnet's performance in classifying 9 distinct error types (CA, CO, UC, CV, HA, OP, FC, MS, CS) for 30 MWP.}
  \label{fig:claude-consistencyExperiment}
  \vspace{-0.5em}
\end{figure*}

To investigate the consistency and accuracy of LLMs in error classification, we choose two high-performance models, GPT-4o and Claude-3.5 sonnet, as error classifiers. We randomly sample 1,000 questions from the GSM8K, AQuA, and MATH datasets. And for each problem, we sample 10 solutions to examine output consistency. The experiment utilizes a consistent prompt shown in Table~\ref{tab:error-classification}. The results in Table~\ref{tab:classification_consistency} indicate that LLM outputs are not invariably uniform for identical problem instances. Furthermore, accuracy in categorizing errors is low, especially when the distinctions between different error types lack clear demarcation. The detailed experimental result is shown in Figure~\ref{fig:GPT-4o-consistencyExperiment} and~\ref{fig:claude-consistencyExperiment}.

We argue that this phenomenon is due to ambiguous error definitions. Vague and overly broad definitions of error types hinder consistent classification. The lack of clear, context-specific criteria leads to variable model judgments due to differences in training data and biases.

We observe that while LLMs perform inconsistently in direct classification tasks, their capability for error analysis is notably strong and highly accurate when explicitly informed that a solution path is incorrect and provided with the correct answer. This finding suggests that although LLMs may exhibit uncertainty and inconsistency in classification tasks, they can accurately identify and explain problems in specific error analysis scenarios. 

To enable effective error identification and analysis, we employ GPT-4o as our error analyzer. An input template, as shown in Figure~\ref{tab:prompt4gptErrorAnalyzer}, guided GPT-4o to not only pinpoint errors but also to provide concise explanations. To validate its performance, we conducted an experiment, sampling 1,000 data points from the MWPES-300k dataset for each (model, dataset) pair. We then manually evaluate GPT-4o's error analysis accuracy, focusing on its error identification precision.

As detailed in Table~\ref{tab:model_accuracy}, experimental results show that as model performance increases, it becomes more challenging for models to analyze complex errors. While high-performance models such as GPT-4o excel in overall accuracy, they sometimes struggle with complex or subtle errors, failing to provide correct analyses. Furthermore, as the dataset difficulty increases, the correction accuracy also tends to decrease, because error correction relies on a thorough analysis of the problem, which may be challenging for high-performance models on some problems. However, it is important to note that the overall error analysis accuracy of the models remains significantly high, suggesting their considerable potential and application value in error analysis tasks.

\subsection{Details of the Experiment Result}
\label{appendixDetailExperimentResult}
In our experiments, we utilize the MWPES-300k dataset, as detailed in Section~\ref{appendixDataConstruction}. Tables~\ref{tab:errors_part1} to~\ref{tab:errors_part7} show the error pattern distribution of 15 LLMs across 39 error types. The error types and their abbreviations are shown in Table~\ref{tab:error_abb}. The results presented in Table ~\ref{tab:errors_part1} to \ref{tab:errors_part7} are based on a single run of the experiment. The \texttt{text-embedding-3-large model} is invoked with default parameter values. Specifically, the \texttt{encoding\_format} parameter is set to float, and the \texttt{dimensions} parameter is set to 3072. 

The KMeans algorithm is applied using the scikit-learn implementation. The following parameter settings are used: \texttt{n\_clusters} is set to a value determined empirically based on the input (e.g., using the elbow method or silhouette analysis); \texttt{init} defaults to \texttt{k-means++}, which intelligently initializes the centroids to accelerate convergence. The \texttt{n\_init} parameter, controlling the number of times the algorithm runs with different centroid seeds, also defaults to \texttt{10}. The maximum number of iterations for each run (\texttt{max\_iter}) defaults to \texttt{300}, and the tolerance for convergence (\texttt{tol}) defaults to \texttt{1e-4}. The \texttt{random\_state} parameter, for reproducibility of the initialization, is set to \texttt{None} (allowing for different results on each run unless a specific seed is desired for consistent behavior). The \texttt{algorithm} parameter implicitly defaults to \texttt{lloyd}. Data is copied using \texttt{copy\_x=True} to allow for safe memory access.

\subsection{Details of Out-of-Distribution Analysis}
\label{appendixDetailOODAnalysis}
The experiments in OOD analysis utilize the TAL-SCQ5K dataset, which is available in both English (TAL-SCQ5K-EN) and Chinese (TAL-SCQ5K-CN) versions. We use the English test set for our experiments. Developed by TAL Education Group, each language version of TAL-SCQ5K comprises 5,000 multiple-choice math questions, divided into 3,000 training examples and 2,000 test examples. The questions span a wide range of mathematical topics, covering primary, junior high, and high school levels. This ensures a diverse and challenging benchmark for evaluating the performance of LLMs on mathematical reasoning tasks. The multiple-choice format allows for straightforward evaluation and comparison of model predictions.

\begin{table*}[h]
\centering
\begin{tabular}{lccc}
\toprule
\textbf{Metric} & \textbf{GPT-4o} & \textbf{Llama-3.1-8B} & \textbf{$\Delta$} \\
\midrule
\multicolumn{4}{l}{\textit{Clustering Quality}} \\
Silhouette Score & 0.71 & 0.64 & -9.9\% \\
Davies-Bouldin Index$^*$ & 0.82 & 1.05 & +28.0\% \\
Calinski-Harabasz Index & 1650.23 & 1412.78 & -14.4\% \\
\midrule
\multicolumn{4}{l}{\textit{Downstream EAP Accuracy (MATH)}} \\
GPT-4o & 79.22\% & 77.15\% & -2.07\% \\
Claude-3.5 & 80.02\% & 76.98\% & -3.04\% \\
Llama-3.1-70b & 78.48\% & 75.23\% & -3.25\% \\
Llama-3.1-8b & 78.23\% & 74.01\% & -4.22\% \\
\bottomrule
\end{tabular}
\caption{Impact of error analyzer quality on clustering metrics and downstream EAP performance. $^*$Lower is better; for all other metrics, higher is better.}
\label{tab:analyzer_ablation}
\end{table*}

\begin{figure*}[h]
  \centering
  \includegraphics[width=0.8\textwidth]{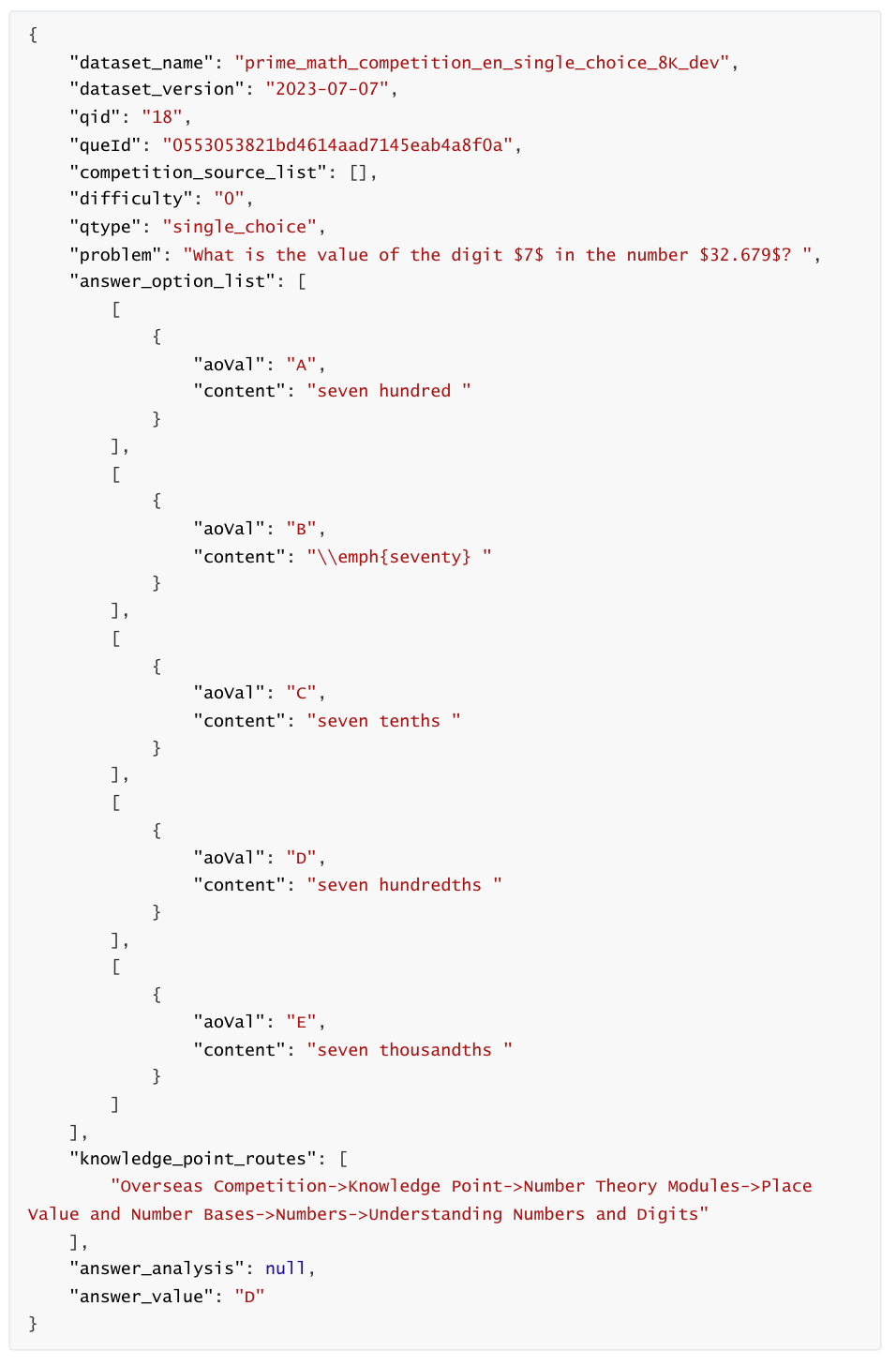}
  \caption{TAL-SCQ5K example}
  \label{fig:TAL-SCQ5KExample}
  \vspace{-0.5em}
\end{figure*}

An example of TAL-SCQ5K dataset is shown in Figure~\ref{fig:TAL-SCQ5KExample}. Each example within the TAL-SCQ5K dataset includes a field named \texttt{knowledge-point-routes}. This field provides a concise description of the specific mathematical concepts and topics relevant to the question. This explicit representation of the knowledge hierarchy is critical for error summary retrieval.

In retrieving relevant error summaries, we employ a cosine similarity algorithm to identify the most related summaries based on the knowledge point route field. The cosine similarity score is a continuous value between 0 and 1, where 1 indicates perfect similarity. In our retrieval process, we have set a similarity threshold of 0.3. This means that only error summaries with a cosine similarity of 0.3 or higher, compared to the input problem's knowledge points, are considered for inclusion in the error-aware prompt. 

To ensure the quality of the information and prevent repetition, we implement a deduplication step: If an error summary has already been added to the prompt, it will not be added again. Furthermore, to avoid overwhelming the model with excessive information and to mitigate potential interference, we restrict the maximum number of included error summaries to 5. This threshold helps to filter out irrelevant error summaries, focusing on the most relevant and helpful error information, while avoiding redundant information and keeping the prompt within reasonable length.

To ensure reliability, the presented results are the average of 5 independent runs.

\subsection{Impact of Alternative Error Analyzers}
\label{appendix:alternative_analyzers}

While our main experiments employ GPT-4o as the error analyzer, we investigate how using alternative models affects the framework's performance. This ablation study quantifies the relationship between analyzer quality and downstream task performance.

\paragraph{Experimental Setup}
We select Llama-3.1-8B as an alternative error analyzer to represent a more accessible but potentially noisier annotator. Using this model, we regenerate error explanations for 1,000 randomly sampled instances from MWPES-300K and rerun our complete framework pipeline.

\paragraph{Results and Analysis}
Table~\ref{tab:analyzer_ablation} presents the comparative analysis between GPT-4o and Llama-3.1-8B as error analyzers. While using the open-source Llama-3.1-8B as an error analyzer does result in performance degradation, the impact remains within acceptable bounds for practical deployment. Clustering metrics show moderate deterioration: the Silhouette Score decreases by 9.9\%, the Davies-Bouldin Index increases by 28\%, and the Calinski-Harabasz Index drops by 14.4\%. Despite these changes, the clustering still maintains sufficient coherence to generate meaningful error categories. More importantly, downstream EAP performance demonstrates remarkable resilience, with accuracy reductions limited to 2\%-4\% across all tested models. Even with Llama-3.1-8B as the analyzer, EAP continues to deliver substantial improvements over baseline methods. This demonstrates that our framework successfully adapts to open-source analyzers, offering a practical pathway for resource-constrained deployments.

\begin{table*}[h!]
\centering
\begin{tabular}{@{}lccccc@{}}
\toprule
Algorithm    & Silhouette & Davies-Bouldin & Calinski-Harabasz & Clu. Num & Calc. Time \\ \midrule
K-Means                 & 0.71              & 0.82                & \textbf{1650.23}                & 39          & \textbf{Fast}       \\
K-Medoids               & \textbf{0.73}              & 0.90                & 1590.59                & 38          & Slow       \\
Hierarchical Clustering & 0.59              & \textbf{0.95}                & 1510.92                & 26          & Very Slow  \\
DBSCAN                  & 0.64              & 1.40                & 1480.01                & 28          & Medium     \\ \bottomrule
\end{tabular}
\caption{Comparison of Clustering Algorithm Performance on the MWPES-300k Dataset.}
\label{tab:clustering_ablation_results}
\end{table*}

\subsection{Ablation Study on Clustering Algorithms for Error Embeddings}
\label{sec:appendix_clustering_ablation}

To evaluate the influence of different clustering algorithms, we conduct an ablation study. The following clustering algorithms are evaluated:
\begin{itemize}
    \item \textbf{K-Means}: A partition-based algorithm.
    \item \textbf{K-Medoids}\citep{kaufman2009finding}: A partition-based algorithm, often more robust to noise and outliers than K-Means.
    \item \textbf{Hierarchical Clustering}\cite{johnson1967hierarchical}: An agglomerative hierarchy-based algorithm.
    \item \textbf{DBSCAN (Density-Based Spatial Clustering of Applications with Noise)}\cite{ester1996density}: A density-based algorithm.
\end{itemize}

All algorithms are applied using their default parameter settings from common library implementations (e.g., scikit-learn) to ensure a standardized comparison. The performance of these algorithms is evaluated on the MWPES-300k dataset using three internal clustering validation metrics:

\begin{itemize}
    \item \textbf{Silhouette Coefficient}: Measures how similar an object is to its own cluster (cohesion) compared to other clusters (separation). Ranges from -1 to 1; a higher value indicates better-defined clusters.
    \item \textbf{Davies-Bouldin Index (DBI)}: Computes the average similarity ratio of each cluster with its most similar cluster. Ranges from 0 to $+\infty$; a lower value indicates better clustering, with clusters being more separated and compact.
    \item \textbf{Calinski-Harabasz Index (CHI)}: Also known as the Variance Ratio Criterion, it is the ratio of the sum of between-cluster dispersion to within-cluster dispersion. Ranges from 0 to $+\infty$; a higher value generally indicates better-defined clusters.
\end{itemize}

The experimental results are shown in Table~\ref{tab:clustering_ablation_results}. The results indicate that K-Medoids achieves the highest Silhouette Coefficient (0.73), suggesting well-separated clusters according to this metric. However, K-Means demonstrates superior performance on both the Davies-Bouldin Index (0.82, lower is better) and the Calinski-Harabasz Index (1650.23, higher is better) compared to K-Medoids (0.90 and 1590.59, respectively). Hierarchical Clustering and DBSCAN generally yield lower scores across these internal validation metrics for this dataset.

A critical factor is computational efficiency. K-Means is notably fast, whereas K-Medoids exhibits a significantly higher computational cost (`slow`). Hierarchical Clustering is `very slow', making it less practical for large datasets like MWPES-300k without specific optimizations. DBSCAN's computation time is moderate.

Considering the balance between clustering quality across multiple intrinsic metrics and computational feasibility, K-Means presents a strong profile. While K-Medoids offers a slight improvement in the Silhouette Coefficient, K-Means performs better on the other two metrics and is considerably more efficient. 

This ablation study reinforces the rationale for using K-Means in our primary analysis due to its favorable trade-off between performance and speed for the scale of our dataset, while also acknowledging K-Medoids as a viable, though more computationally intensive.

\subsection{Per-Error Type Impact of Error-Aware Prompting}
\label{sec:per-error-type-analysis}
To gain a more granular understanding of the efficacy of Error-Aware Prompting, beyond overall accuracy metrics, we analyze its impact on specific error types. This investigation aims to identify which categories of errors are more readily corrected by the error summary feedback and, conversely, which types might persist or even be exacerbated.

\paragraph{Experimental Setup}
For this detailed analysis, we select a representative set of error types from the 39 distinct categories identified in our study. For each chosen error type, 500 samples exhibiting that specific error are subjected to both the baseline model and the Error-Aware Prompting approach. The accuracy, defined as the percentage of correctly resolved instances, is then compared.

\paragraph{Results and Discussion}

The results reveal distinct patterns in how Error-Aware Prompting affects different error categories:

\begin{itemize}
    \item \textbf{Limited Improvement or Worsening for Deep Comprehension Errors}: For error types that stem from a fundamental misunderstanding of the problem's requirements or complex constraints, Error-Aware Prompting shows minimal positive impact. Notably, for \textit{"Misunderstanding of problem requirements (MOP),"} a slight decrease in accuracy (-1.40\%) is observed. Similarly, \textit{"Insufficient understanding/consideration of problem constraints (IUC)"} sees only a modest improvement (3.70\%). This suggests that high-level conceptual errors are less amenable to correction through the algorithm.

    \item \textbf{Significant Improvement for Specific Operational Errors}: In contrast, Error-Aware Prompting yields substantial accuracy gains for errors related to specific calculation procedures or the direct application of formulas. For instance, remarkable improvements are seen for \textit{"Failure to accurately apply speed formulas and conversions (FAC)"} (+13.50 \%) and \textit{"Incorrect calculation of the least common multiple (LCM)"} (+14.50 \%). Other operational errors like \textit{"Miscalculation during the algebraic manipulation (MAM)"} (+5.35 \%) and \textit{"Unit Error (UNE)"} (+6.50 \%) also benefites significantly.
\end{itemize}

Table~\ref{tab:per_error_accuracy} presents the accuracy changes for a selection of error types. Due to the extensive number of error categories (39 in total), this table showcases a representative subset to illustrate the observed results.

\subsection{Comparison with Static Error Example Baseline}
\label{appendix:static_baseline}

To further validate the effectiveness of our dynamic error-aware approach, we implement and evaluate a static few-shot baseline that incorporates error examples directly into the prompt. This baseline addresses whether simply exposing models to common error patterns through static examples could achieve similar benefits to our dynamic retrieval mechanism.

\begin{lstlisting}[basicstyle=\small\ttfamily, breaklines=true]
Question: [Example Problem 1]
Correct: [Proper reasoning and solution]
Incorrect: [Erroneous reasoning with common mistake]

[... additional examples ...]

Question: [Test Problem]
Correct: 
\end{lstlisting}

\paragraph{Experimental Design}
We construct a static few-shot prompt (CoT-Error) that provides the model with paired examples of correct and incorrect reasoning paths. Each example triplet consists of: (1) a mathematical problem, (2) its correct solution with proper reasoning, and (3) an erroneous solution demonstrating a common error type identified in our analysis. We select five representative error types from our 39 categories, focusing on the most frequent errors: misunderstanding problem requirements, algebraic manipulation errors, unit conversion mistakes, incomplete constraint consideration, and calculation errors.

\paragraph{Results and Analysis}
Table~\ref{tab:static_baseline} presents the comparative performance on the MATH dataset using Llama-3.1-8B. Surprisingly, the static error example baseline (CoT-Error) not only fails to improve upon standard CoT but actually degrades performance by 4.51 percentage points.

\begin{table}[t]
\centering
\begin{tabular}{lc}
\toprule
\textbf{Method} & \textbf{Accuracy (\%)} \\
\midrule
CoT (Baseline) & 75.76 \\
CoT-Error (Static Examples) & 71.25 \\
PoT & 73.43 \\
Complex-CoT & 81.34 \\
CoT (Self-Consistency) & 83.87 \\
\midrule
\textbf{EAP (Dynamic Retrieval)} & \textbf{80.23} \\
\bottomrule
\end{tabular}
\caption{Performance comparison of static error examples versus dynamic error-aware prompting on MATH dataset with Llama-3.1-8B.}
\label{tab:static_baseline}
\end{table}

This counterintuitive result reveals a fundamental limitation of static error guidance: the lack of contextual relevance. Static examples, regardless of their quality, cannot adapt to the specific mathematical concepts, problem structure, or potential pitfalls of each individual question. In contrast, our dynamic EAP approach retrieves error patterns specifically relevant to each problem's knowledge requirements, achieving 80.23\% accuracy—an improvement of 4.47\% over standard CoT and 8.98\% over the static error baseline.

The performance degradation in CoT-Error likely stems from cognitive interference: irrelevant error examples may prime the model toward mistakes it would not naturally make, while failing to address the actual challenges present in the test problem. This finding underscores the critical importance of our dynamic retrieval mechanism and validates our design choice to adaptively select error guidance based on problem-specific characteristics rather than relying on fixed examples.

\begin{table*}[h!]
\centering
\caption{Accuracy Comparison by Error Type for Baseline vs. Error-Aware Prompting (EAP). Accuracies are percentages. Accuracy Change is the absolute percentage point difference.}
\label{tab:per_error_accuracy}
\resizebox{\textwidth}{!}{% Resize table to fit within text width
\begin{tabular}{@{}llccc@{}}
\toprule
Error Acronym & Description                                           & Baseline Acc. (\%) & EAP Acc. (\%) & Acc. Change (\% pts) \\ \midrule
MOP           & Misunderstanding of problem requirements                & 55.20              & 53.80           & -1.40                  \\
LOV           & Lack of verification for final answer                   & 62.50              & 63.10           & 0.60                   \\
MAM           & Miscalculation during the algebraic manipulation        & 70.15              & 75.50           & 5.35                   \\
IUC           & Insufficient understanding/consideration of constraints & 68.30              & 72.00           & 3.70                   \\
LAB           & Lacks thorough analysis of boundary conditions          & 75.80              & 81.20           & 5.40                   \\
UNE           & Unit Error                                              & 82.00              & 88.50           & 6.50                   \\
AIO           & Assumed independence of overlapping events              & 78.90              & 85.00           & 6.10                   \\
MPI           & Misapplication of probability formula (independent events) & 85.60              & 92.10           & 6.50                   \\
FAC           & Failure to accurately apply speed formulas/conversions  & 72.40              & 85.90           & 13.50                  \\
LCM           & Incorrect calculation of the least common multiple (LCM) & 65.70              & 80.20           & 14.50                  \\
IAF           & Inconsistent application of formulas                    & 73.50              & 78.10           & 4.60                   \\
LOS           & Lack of Simplification                                  & 80.00              & 86.20           & 6.20                   \\ \bottomrule
\end{tabular}%
}
\end{table*}


\begin{table*}
\centering
\resizebox{0.95\linewidth}{!}{%
\begin{tblr}{
  width = \linewidth,
  colspec = {Q[871]Q[73]},
  column{2} = {c},
  hline{1,41} = {-}{0.08em},
  hline{2} = {-}{},
}
\textbf{Error Type} & \textbf{Code}\\
Misunderstanding of problem requirements & MOP\\
Lack of verification for final answer. & LOV\\
Miscalculation during the algebraic manipulation. & MAM\\
Insufficient understanding/consideration of problem constraints. & IUC\\
Ambiguous problem parameters. & APP\\
Misinterpreted conditions led to incorrect assumptions. & MCI\\
Incorrect application of mathematical formulas/concepts. & IAM\\
Inconsistent application of formulas. & IAF\\
Lack of logical reasoning in arriving at the answer. & LLR\\
Inconsistent Variable Substitutions & IVS\\
Misinterpretation of what the problem is asking for. & MWP\\
Lack of Simplification. & LOS\\
Incorrect equation setup due to misinterpreted conditions. & IES\\
Lacks thorough analysis of boundary conditions & LAB\\
Overcomplication by introducing irrelevant elements resulted in the wrong expression. & ORI\\
The notation used led to confusion in mathematical operations. & NCM\\
Misplaced focus on solving an unnecessary/unsolvable variable. & MFS\\
Miscalculation during the equation solving process. & MES\\
Unit Error. & UNE\\
Misinterpretation of rounding rules. & MRR\\
Irrelevant Content. & IC\\
Failure to consider all possible solutions. & FCS\\
Overreliance on assumptions instead of analysis. & ORA\\
Incorrect factorization/Confusion between different forms of number factorizations. & IFC\\
Incorrect calculation during simplification steps. & ICS\\
Improper application of multiplication relationships & IMR\\
Incorrect application of combinatorial principles. & ICP\\
Assumed independence of overlapping events. & AIO\\
Misapplication of the probability formula for independent events. & MPI\\
Failure to accurately apply speed formulas and conversions. & FAC\\
Misinterpretation of geometric relationships in the problem context. & MGR\\
Misapplication of perimeter and area formulas for rectangles. & MPA\\
Misinterpretation of scaling language led to errors in calculations. & MSC\\
Incorrectly assumed equivalence of different algebraic expressions. & IAE\\
Incorrect calculation of the least common multiple (LCM). & ILC\\
Misunderstanding of expected answer format. & MEF\\
Failure to distinguish between selling price and cost price. & FSP\\
Incorrect application of trigonometric functions. & ITF\\
Misapplication of modular reasoning. & MMR
\end{tblr}
}
\caption{Error Types and Their Abbreviations}
\label{tab:error_abb}
\end{table*}

\setlength{\tabcolsep}{4pt}
\begin{table*}[htbp]
\centering
\small
\begin{tabular}{lcccccc}
\toprule
Model & MOP & LOV & MAM & IUC & APP & MCI \\
\midrule
gpt-o1-mini & 1699[4.83\%] & 2368[6.73\%] & 1769[5.03\%] & 2535[7.20\%] & 0[0.00\%] & 1630[4.63\%] \\
gpt-4o & 5976[10.31\%] & 5781[9.97\%] & 3970[6.85\%] & 4026[6.94\%] & 0[0.00\%] & 3482[6.01\%] \\
gpt-3.5 turbo & 10907[11.93\%] & 10643[11.64\%] & 8121[8.88\%] & 5656[6.19\%] & 0[0.00\%] & 5335[5.83\%] \\
gemma-2-27b-it & 7258[10.56\%] & 7843[11.41\%] & 7996[11.63\%] & 5697[8.29\%] & 0[0.00\%] & 4054[5.90\%] \\
gemma-2-9b-it & 12063[13.57\%] & 10893[12.25\%] & 8818[9.92\%] & 8316[9.35\%] & 933[1.05\%] & 4959[5.58\%] \\
claude-3.5-sonnet & 6366[9.99\%] & 5391[8.46\%] & 3789[5.95\%] & 4541[7.13\%] & 0[0.00\%] & 4040[6.34\%] \\
claude-3.5-haiku & 11353[11.29\%] & 8483[8.44\%] & 7717[7.67\%] & 7884[7.84\%] & 1908[1.90\%] & 4151[4.13\%] \\
llama-3.1-405b & 5614[8.58\%] & 8024[12.27\%] & 3928[6.01\%] & 6171[9.44\%] & 0[0.00\%] & 3664[5.60\%] \\
llama-3.1-70b & 12105[12.22\%] & 11339[11.45\%] & 8901[8.99\%] & 10378[10.48\%] & 1100[1.11\%] & 7550[7.62\%] \\
llama-3.1-8b & 17315[13.56\%] & 13944[10.92\%] & 10489[8.21\%] & 8929[6.99\%] & 4792[3.75\%] & 10517[8.24\%] \\
Qwen2.5-72b-Instruct & 5600[8.06\%] & 7174[10.32\%] & 6756[9.72\%] & 5461[7.86\%] & 0[0.00\%] & 4137[5.95\%] \\
Qwen2.5-7b-Instruct & 9375[10.91\%] & 5544[6.45\%] & 7397[8.61\%] & 3106[3.62\%] & 1212[1.41\%] & 7076[8.24\%] \\
Qwen2-7b-Instruct & 16995[14.57\%] & 10559[9.05\%] & 13178[11.30\%] & 9890[8.48\%] & 3789[3.25\%] & 7982[6.84\%] \\
glm-4-9b-chat & 21160[14.20\%] & 13707[9.20\%] & 11673[7.83\%] & 8985[6.03\%] & 3984[2.67\%] & 11311[7.59\%] \\
chatglm3-6b & 27261[14.60\%] & 21508[11.52\%] & 18555[9.93\%] & 8957[4.80\%] & 5293[2.83\%] & 10838[5.80\%] \\
\bottomrule
\end{tabular}
\caption{Error Metrics (Part 1)}
\label{tab:errors_part1}
\end{table*}

\begin{table*}[htbp]
\centering
\small
\begin{tabular}{lcccccc}
\toprule
Model & IAM & IAF & LLR & IVS & MWP & LOS \\
\midrule
gpt-o1-mini & 3232[9.18\%] & 0[0.00\%] & 0[0.00\%] & 0[0.00\%] & 0[0.00\%] & 0[0.00\%] \\
gpt-4o & 5266[9.08\%] & 0[0.00\%] & 0[0.00\%] & 0[0.00\%] & 0[0.00\%] & 1212[2.09\%] \\
gpt-3.5 turbo & 4848[5.30\%] & 0[0.00\%] & 0[0.00\%] & 0[0.00\%] & 836[0.91\%] & 1059[1.16\%] \\
gemma-2-27b-it & 5558[8.08\%] & 0[0.00\%] & 0[0.00\%] & 0[0.00\%] & 0[0.00\%] & 947[1.38\%] \\
gemma-2-9b-it & 8929[10.04\%] & 0[0.00\%] & 599[0.67\%] & 432[0.49\%] & 1059[1.19\%] & 2285[2.57\%] \\
claude-3.5-sonnet & 4611[7.23\%] & 0[0.00\%] & 0[0.00\%] & 0[0.00\%] & 0[0.00\%] & 0[0.00\%] \\
claude-3.5-haiku & 6951[6.91\%] & 0[0.00\%] & 780[0.78\%] & 0[0.00\%] & 2647[2.63\%] & 2967[2.95\%] \\
llama-3.1-405b & 3789[5.79\%] & 0[0.00\%] & 0[0.00\%] & 0[0.00\%] & 0[0.00\%] & 641[0.98\%] \\
llama-3.1-70b & 8609[8.69\%] & 404[0.41\%] & 0[0.00\%] & 0[0.00\%] & 1546[1.56\%] & 975[0.98\%] \\
llama-3.1-8b & 6965[5.45\%] & 1351[1.06\%] & 2647[2.07\%] & 850[0.67\%] & 3566[2.79\%] & 0[0.00\%] \\
Qwen2.5-72b-Instruct & 3900[5.61\%] & 0[0.00\%] & 0[0.00\%] & 0[0.00\%] & 0[0.00\%] & 1156[1.66\%] \\
Qwen2.5-7b-Instruct & 7355[8.56\%] & 933[1.09\%] & 933[1.09\%] & 961[1.12\%] & 2897[3.37\%] & 390[0.45\%] \\
Qwen2-7b-Instruct & 6213[5.33\%] & 1867[1.60\%] & 2507[2.15\%] & 1351[1.16\%] & 1894[1.62\%] & 1407[1.21\%] \\
glm-4-9b-chat & 8525[5.72\%] & 2786[1.87\%] & 5753[3.86\%] & 2048[1.37\%] & 8274[5.55\%] & 0[0.00\%] \\
chatglm3-6b & 12189[6.53\%] & 3106[1.66\%] & 11687[6.26\%] & 4736[2.54\%] & 8832[4.73\%] & 1379[0.74\%] \\
\bottomrule
\end{tabular}
\caption{Error Metrics (Part 2)}
\label{tab:errors_part2}
\end{table*}

\begin{table*}[htbp]
\centering
\small
\begin{tabular}{lcccccc}
\toprule
Model & IES & LAB & ORI & NCM & MFS & MES \\
\midrule
gpt-o1-mini & 1686[4.79\%] & 2326[6.61\%] & 1867[5.30\%] & 0[0.00\%] & 460[1.31\%] & 557[1.58\%] \\
gpt-4o & 2675[4.61\%] & 1351[2.33\%] & 1323[2.28\%] & 0[0.00\%] & 0[0.00\%] & 1073[1.85\%] \\
gpt-3.5 turbo & 5544[6.06\%] & 1087[1.19\%] & 905[0.99\%] & 0[0.00\%] & 766[0.84\%] & 5878[6.43\%] \\
gemma-2-27b-it & 3775[5.49\%] & 1365[1.99\%] & 780[1.13\%] & 0[0.00\%] & 0[0.00\%] & 1142[1.66\%] \\
gemma-2-9b-it & 5447[6.13\%] & 0[0.00\%] & 0[0.00\%] & 0[0.00\%] & 1226[1.38\%] & 1894[2.13\%] \\
claude-3.5-sonnet & 3371[5.29\%] & 1616[2.54\%] & 1546[2.43\%] & 0[0.00\%] & 0[0.00\%] & 3274[5.14\%] \\
claude-3.5-haiku & 5112[5.08\%] & 0[0.00\%] & 0[0.00\%] & 0[0.00\%] & 2326[2.31\%] & 4764[4.74\%] \\
llama-3.1-405b & 3524[5.39\%] & 1867[2.85\%] & 1240[1.90\%] & 0[0.00\%] & 0[0.00\%] & 2187[3.34\%] \\
llama-3.1-70b & 4402[4.45\%] & 0[0.00\%] & 0[0.00\%] & 696[0.70\%] & 0[0.00\%] & 3287[3.32\%] \\
llama-3.1-8b & 4876[3.82\%] & 0[0.00\%] & 1170[0.92\%] & 1644[1.29\%] & 1867[1.46\%] & 4555[3.57\%] \\
Qwen2.5-72b-Instruct & 2800[4.03\%] & 892[1.28\%] & 1630[2.34\%] & 0[0.00\%] & 0[0.00\%] & 1240[1.78\%] \\
Qwen2.5-7b-Instruct & 2856[3.32\%] & 0[0.00\%] & 209[0.24\%] & 0[0.00\%] & 1546[1.80\%] & 2090[2.43\%] \\
Qwen2-7b-Instruct & 4499[3.86\%] & 0[0.00\%] & 0[0.00\%] & 1073[0.92\%] & 1184[1.02\%] & 1393[1.19\%] \\
glm-4-9b-chat & 4876[3.27\%] & 0[0.00\%] & 0[0.00\%] & 1365[0.92\%] & 0[0.00\%] & 7411[4.97\%] \\
chatglm3-6b & 5224[2.80\%] & 0[0.00\%] & 0[0.00\%] & 1811[0.97\%] & 0[0.00\%] & 3552[1.90\%] \\
\bottomrule
\end{tabular}
\caption{Error Metrics (Part 3)}
\label{tab:errors_part3}
\end{table*}

\begin{table*}[htbp]
\centering
\small
\begin{tabular}{lcccccc}
\toprule
Model & UNE & MRR & IC & FCS & ORA & IFC \\
\midrule
gpt-o1-mini & 0[0.00\%] & 320[0.91\%] & 0[0.00\%] & 1992[5.66\%] & 557[1.58\%] & 404[1.15\%] \\
gpt-4o & 0[0.00\%] & 1365[2.35\%] & 0[0.00\%] & 1351[2.33\%] & 3552[6.13\%] & 1574[2.71\%] \\
gpt-3.5 turbo & 822[0.90\%] & 2758[3.02\%] & 0[0.00\%] & 2563[2.80\%] & 3204[3.50\%] & 1031[1.13\%] \\
gemma-2-27b-it & 0[0.00\%] & 989[1.44\%] & 0[0.00\%] & 1853[2.69\%] & 2326[3.38\%] & 446[0.65\%] \\
gemma-2-9b-it & 0[0.00\%] & 794[0.89\%] & 0[0.00\%] & 0[0.00\%] & 2995[3.37\%] & 1379[1.55\%] \\
claude-3.5-sonnet & 0[0.00\%] & 2006[3.15\%] & 0[0.00\%] & 1964[3.08\%] & 975[1.53\%] & 1114[1.75\%] \\
claude-3.5-haiku & 1853[1.84\%] & 1337[1.33\%] & 0[0.00\%] & 878[0.87\%] & 2034[2.02\%] & 0[0.00\%] \\
llama-3.1-405b & 0[0.00\%] & 3831[5.86\%] & 1240[1.90\%] & 2076[3.17\%] & 2048[3.13\%] & 599[0.92\%] \\
llama-3.1-70b & 0[0.00\%] & 2507[2.53\%] & 1867[1.88\%] & 975[0.98\%] & 1059[1.07\%] & 0[0.00\%] \\
llama-3.1-8b & 1240[0.97\%] & 3733[2.92\%] & 5558[4.35\%] & 0[0.00\%] & 2298[1.80\%] & 0[0.00\%] \\
Qwen2.5-72b-Instruct & 460[0.66\%] & 2382[3.43\%] & 0[0.00\%] & 2368[3.41\%] & 1797[2.59\%] & 0[0.00\%] \\
Qwen2.5-7b-Instruct & 1059[1.23\%] & 2382[2.77\%] & 0[0.00\%] & 474[0.55\%] & 0[0.00\%] & 752[0.88\%] \\
Qwen2-7b-Instruct & 1853[1.59\%] & 3106[2.66\%] & 2382[2.04\%] & 0[0.00\%] & 1825[1.56\%] & 0[0.00\%] \\
glm-4-9b-chat & 2145[1.44\%] & 2716[1.82\%] & 6101[4.09\%] & 0[0.00\%] & 0[0.00\%] & 0[0.00\%] \\
chatglm3-6b & 2257[1.21\%] & 0[0.00\%] & 7968[4.27\%] & 0[0.00\%] & 0[0.00\%] & 0[0.00\%] \\
\bottomrule
\end{tabular}
\caption{Error Metrics (Part 4)}
\label{tab:errors_part4}
\end{table*}

\begin{table*}[htbp]
\centering
\small
\begin{tabular}{lcccccc}
\toprule
Model & ICS & IMR & ICP & AIO & MPI & FAC \\
\midrule
gpt-o1-mini & 850[2.41\%] & 0[0.00\%] & 1532[4.35\%] & 2076[5.90\%] & 1365[3.88\%] & 0[0.00\%] \\
gpt-4o & 1630[2.81\%] & 0[0.00\%] & 2688[4.64\%] & 2466[4.25\%] & 1811[3.12\%] & 0[0.00\%] \\
gpt-3.5 turbo & 3775[4.13\%] & 0[0.00\%] & 2173[2.38\%] & 2020[2.21\%] & 947[1.04\%] & 1087[1.19\%] \\
gemma-2-27b-it & 543[0.79\%] & 0[0.00\%] & 3873[5.63\%] & 1309[1.90\%] & 836[1.22\%] & 1295[1.88\%] \\
gemma-2-9b-it & 1073[1.21\%] & 0[0.00\%] & 4430[4.98\%] & 1950[2.19\%] & 515[0.58\%] & 683[0.77\%] \\
claude-3.5-sonnet & 1811[2.84\%] & 0[0.00\%] & 1908[2.99\%] & 1003[1.57\%] & 1240[1.95\%] & 1170[1.84\%] \\
claude-3.5-haiku & 2897[2.88\%] & 0[0.00\%] & 4290[4.27\%] & 1073[1.07\%] & 599[0.60\%] & 4806[4.78\%] \\
llama-3.1-405b & 1087[1.66\%] & 0[0.00\%] & 2549[3.90\%] & 752[1.15\%] & 1212[1.85\%] & 1254[1.92\%] \\
llama-3.1-70b & 1532[1.55\%] & 0[0.00\%] & 4597[4.64\%] & 864[0.87\%] & 613[0.62\%] & 1699[1.72\%] \\
llama-3.1-8b & 1365[1.07\%] & 1867[1.46\%] & 0[0.00\%] & 0[0.00\%] & 0[0.00\%] & 3120[2.44\%] \\
Qwen2.5-72b-Instruct & 1435[2.06\%] & 0[0.00\%] & 3399[4.89\%] & 1114[1.60\%] & 1254[1.80\%] & 1588[2.28\%] \\
Qwen2.5-7b-Instruct & 1908[2.22\%] & 724[0.84\%] & 6631[7.72\%] & 0[0.00\%] & 0[0.00\%] & 3761[4.38\%] \\
Qwen2-7b-Instruct & 1811[1.55\%] & 1588[1.36\%] & 2967[2.54\%] & 0[0.00\%] & 0[0.00\%] & 947[0.81\%] \\
glm-4-9b-chat & 2549[1.71\%] & 4040[2.71\%] & 1532[1.03\%] & 0[0.00\%] & 0[0.00\%] & 4862[3.26\%] \\
chatglm3-6b & 1825[0.98\%] & 4764[2.55\%] & 1087[0.58\%] & 0[0.00\%] & 0[0.00\%] & 4569[2.45\%] \\
\bottomrule
\end{tabular}
\caption{Error Metrics (Part 5)}
\label{tab:errors_part5}
\end{table*}

\begin{table*}[htbp]
\centering
\small
\begin{tabular}{lcccccc}
\toprule
Model & MGR & MPA & MSC & IAE & ILC & MEF \\
\midrule
gpt-o1-mini & 1853[5.26\%] & 0[0.00\%] & 0[0.00\%] & 0[0.00\%] & 515[1.46\%] & 0[0.00\%] \\
gpt-4o & 1727[2.98\%] & 0[0.00\%] & 0[0.00\%] & 0[0.00\%] & 878[1.51\%] & 0[0.00\%] \\
gpt-3.5 turbo & 2647[2.89\%] & 0[0.00\%] & 947[1.04\%] & 599[0.66\%] & 1532[1.68\%] & 1379[1.51\%] \\
gemma-2-27b-it & 2493[3.63\%] & 794[1.15\%] & 557[0.81\%] & 933[1.36\%] & 961[1.40\%] & 0[0.00\%] \\
gemma-2-9b-it & 2410[2.71\%] & 1686[1.90\%] & 1365[1.54\%] & 0[0.00\%] & 432[0.49\%] & 0[0.00\%] \\
claude-3.5-sonnet & 3371[5.29\%] & 0[0.00\%] & 0[0.00\%] & 2145[3.37\%] & 1713[2.69\%] & 2076[3.26\%] \\
claude-3.5-haiku & 2424[2.41\%] & 2173[2.16\%] & 2702[2.69\%] & 0[0.00\%] & 1867[1.86\%] & 1532[1.52\%] \\
llama-3.1-405b & 2716[4.15\%] & 710[1.09\%] & 0[0.00\%] & 460[0.70\%] & 975[1.49\%] & 683[1.04\%] \\
llama-3.1-70b & 1797[1.81\%] & 1825[1.84\%] & 1727[1.74\%] & 0[0.00\%] & 2939[2.97\%] & 334[0.34\%] \\
llama-3.1-8b & 975[0.76\%] & 2981[2.33\%] & 3970[3.11\%] & 0[0.00\%] & 738[0.58\%] & 1686[1.32\%] \\
Qwen2.5-72b-Instruct & 1309[1.88\%] & 1045[1.50\%] & 1114[1.60\%] & 1393[2.00\%] & 1672[2.40\%] & 0[0.00\%] \\
Qwen2.5-7b-Instruct & 2159[2.51\%] & 3134[3.65\%] & 1727[2.01\%] & 1797[2.09\%] & 850[0.99\%] & 1658[1.93\%] \\
Qwen2-7b-Instruct & 3232[2.77\%] & 3343[2.87\%] & 2535[2.17\%] & 0[0.00\%] & 696[0.60\%] & 1295[1.11\%] \\
glm-4-9b-chat & 1128[0.76\%] & 0[0.00\%] & 5015[3.36\%] & 0[0.00\%] & 306[0.21\%] & 4499[3.02\%] \\
chatglm3-6b & 794[0.43\%] & 5084[2.72\%] & 3343[1.79\%] & 0[0.00\%] & 446[0.24\%] & 6854[3.67\%] \\
\bottomrule
\end{tabular}
\caption{Error Metrics (Part 6)}
\label{tab:errors_part6}
\end{table*}

\begin{table*}[htbp]
\centering
\small
\begin{tabular}{lccc}
\toprule
Model & FSP & ITF & MMR \\
\midrule
gpt-o1-mini & 0[0.00\%] & 2034[5.78\%] & 1574[4.47\%] \\
gpt-4o & 0[0.00\%] & 1658[2.86\%] & 1156[1.99\%] \\
gpt-3.5 turbo & 0[0.00\%] & 1128[1.23\%] & 1240[1.36\%] \\
gemma-2-27b-it & 961[1.40\%] & 1240[1.80\%] & 933[1.36\%] \\
gemma-2-9b-it & 1351[1.52\%] & 0[0.00\%] & 0[0.00\%] \\
claude-3.5-sonnet & 0[0.00\%] & 1755[2.75\%] & 933[1.46\%] \\
claude-3.5-haiku & 3051[3.03\%] & 0[0.00\%] & 0[0.00\%] \\
llama-3.1-405b & 0[0.00\%] & 1435[2.19\%] & 1128[1.73\%] \\
llama-3.1-70b & 2856[2.88\%] & 543[0.55\%] & 0[0.00\%] \\
llama-3.1-8b & 2688[2.11\%] & 0[0.00\%] & 0[0.00\%] \\
Qwen2.5-72b-Instruct & 3873[5.57\%] & 1783[2.57\%] & 780[1.12\%] \\
Qwen2.5-7b-Instruct & 2382[2.77\%] & 627[0.73\%] & 0[0.00\%] \\
Qwen2-7b-Instruct & 3274[2.81\%] & 0[0.00\%] & 0[0.00\%] \\
glm-4-9b-chat & 2285[1.53\%] & 0[0.00\%] & 0[0.00\%] \\
chatglm3-6b & 2856[1.53\%] & 0[0.00\%] & 0[0.00\%] \\
\bottomrule
\end{tabular}
\caption{Error Metrics (Part 7)}
\label{tab:errors_part7}
\end{table*}

\begin{table*}
\centering
\begin{tblr}{
  width = \linewidth,
  colspec = {Q[180]Q[80]Q[80]Q[80]Q[80]Q[80]Q[80]},
  column{even} = {c},
  hline{1,17} = {-}{0.08em},
  hline{2} = {-}{0.05em},
}
\textbf{Model Name} & \textbf{Launch} & \textbf{Parameters} & \textbf{MATH} & \textbf{AQuA} & \textbf{SVAMP} & \textbf{GSM8K} \\
GPT-4o                    & 2024 & / & 88\% & 92\% & 98\%  & 96\%  \\
o1-mini                   & 2024 & / & 85\% & 89\% & 97\%  & 99\%  \\
GPT-3.5-turbo             & 2022 & / & 91\% & 93\% & 96\%  & 97\%  \\
Llama-3.1-8B              & 2024 & 8B   & /    & /    & 97\%  & 96\%  \\
Llama-3.1-70B             & 2024 & 70B  & 93\% & 94\% & 98\%  & 95\%  \\
Llama-3.1-405B            & 2024 & 405B & 89\% & 92\% & 97\%  & 97\%  \\
glm-4-9b-chat             & 2024 & 9B   & /    & /    & 96\%  & 98\%  \\
Chatglm3-6b               & 2024 & 6B   & /    & /    & 97\%  & 98\%  \\
Qwen2-7B-Instruct         & 2024 & 7B   & /    & /    & 98\%  & 96\%  \\
Qwen2.5-7B-Instruct       & 2024 & 7B   & /    & /    & 97\%  & 97\%  \\
Qwen2.5-72B-Instruct      & 2024 & 72B  & 91\% & 90\% & 98\%  & 96\%  \\
Gemma-2-9b-it             & 2024 & 9B   & 94\% & 95\% & 97\%  & 97\%  \\
Gemma-2-27b-it            & 2024 & 27B  & 88\% & 91\% & 97\%  & 98\%  \\
Claude-3.5 Sonnet         & 2024 & /    & 91\% & 90\% & 98\%  & 97\%  \\
Claude-3 Haiku            & 2024 & /    & 93\% & 95\% & 98\%  & 96\%  
\end{tblr}
\caption{Error Analysis Accuracy of Various Models on Different Datasets}
\label{tab:model_accuracy}
\end{table*}

\begingroup

\begin{table*}
    \centering
    \small
    \renewcommand{\arraystretch}{0.95}  % 减小行间距
    \begin{tabular}{p{\linewidth}}
        \toprule
        \textbf{Instruction:} You are an expert in reviewing math problem-solving steps. Your specialty is identifying errors in provided solution steps and summarizing them into abstract categories.You will receive information containing the original question, the solution steps and the correct answer.\\
        1. Provide a concise explanation of the errors in the problem-solving process. If the response correctly solves the problem, output "Solution is right."\\
        2. After the detailed explanation, please briefly summarize the core reasons for the mistake into abstract categories, avoiding details related to the specific problem. The abstract reasons should be listed with numbered points, and each line should not exceed 15 words. Try to minimize the number of reasons (less than 4).\\
        The output should clearly separate the detailed error analysis from the summarized reasons. 
        \\
        \midrule
        Q: Uriah's book bag is getting too heavy for him. He needs to remove 15 pounds from it. His comic books weigh 1/4 pound each and his toys weigh 1/2 pound each. If he removes 30 comic books, how many toys does he need to remove?\\
        R: Each comic book weighs 1/4 pound and Uriah needs to remove 15 pounds in total. So the number of comic books he needs to remove is 15 / (1/4) = 15 * 4 = 60 comic books. Since each toy weighs 1/2 pound, the number of toys you need to remove is 60 toys / (1/2) = 60 * 2 = 120 toys. So the answer is 120.\\
        A: 15\\
        Analysis: The solution contains errors in the problem-solving process: \\1. Misinterpretation of the Problem: The response incorrectly calculates the number of comic books to be removed. The problem states that Uriah removes 30 comic books, not that he needs to calculate how many to remove. \\2. Incorrect Calculation for Toys: The response incorrectly calculates the number of toys to remove. After removing 30 comic books, the weight removed is ($30 \times \frac{1}{4} = 7.5$) pounds. Uriah still needs to remove \\($15 - 7.5 = 7.5$) pounds.\\3. Correct Calculation for Toys: To remove the remaining 7.5 pounds using toys that weigh 1/2 pound each, the number of toys to remove is ($7.5 \div \frac{1}{2} = 15$) toys.\\The correct answer is 15 toys, not 120.
        \\
        \midrule
        Q: I have 10 liters of orange drink that are two-thirds water and I wish to add it to 15 liters of pineapple drink that is three-fifths water. But as I pour it, I spill one liter of the orange drink. How much water is in the remaining 24 liters?\\
        R: Each comic book weighs 1/4 pound and Uriah needs to remove 15 pounds in total. So the number of comic books he needs to remove is 15 / (1/4) = 15 * 4 = 60 comic books. Since each toy weighs 1/2 pound, the number of toys you need to remove is 60 toys / (1/2) = 60 * 2 = 120 toys. So the answer is 120.
        The shore was 64 seconds away, so Julia rowed towards shore for a total of 64/0.8 = 80 units of 10 feet. 
        For every 10 feet she rowed towards shore, the boat took on 2 liters of water. 
        Therefore, the boat took on a total of 80*2 = 160 liters of water by the time Julia reached shore.
        \\
        A: 15\\
        Analysis: The solution contains multiple errors in its calculations:\\1. Misinterpretation of the spill: When $1$ liter of orange drink is spilled, it affects the amount of water contributed by the orange drink proportionately. The solution incorrectly subtracts 1 directly from the total water content, rather than calculating the water in the spilled 1 liter of orange drink and adjusting accordingly.\\2. Incorrect total water computation: The orange drink is two-thirds water, so spilling 1 liter of it means ( $\frac{2}{3} \times 1 = \frac{2}{3} $) liters of water is lost. The correct adjustment requires subtracting ( $\frac{2}{3}$) liters of water from the total, not 1 liter as stated in the response.\\3. Final calculation discrepancy: The total water content in the remaining mixture should be accurately recomputed after taking the spill into account. The response's claim of $\frac{56}{3}$ liters of water is incorrect. The correct approach follows these steps:- Before the spill:  - Water in orange drink: ( $\frac{2}{3} \times 10 = \frac{20}{3}$) liters. - Water in pineapple drink: $\frac{3}{5} \times 15 = 9$ liters.  - Total water in drinks: $\frac{20}{3} + 9 = \frac{47}{3}$ liters.- After the spill:  - Water lost from the spill: ( $\frac{2}{3}$ ) liters.  - Remaining water:  $\frac{47}{3} - \frac{2}{3} = \frac{45}{3} = 15 $ liters. Thus, there is \( 15 \) liters of water in the remaining \(24\) liters of the mixture.\\The provided solution has calculation errors in considering the impact of the spill and in the final total. The correct answer is \(15\) liters, not ( $\frac{56}{3}$).
        \\
        \midrule
        Q: Simplify $-3(1+4i)+i(-2-i)$.\\
        R: Step 1: Simplify the first term of the expression. The first term is $-3(1+4i)$. Using the distributive property, we multiply $-3$ by both $1$ and $4i$, giving $-3 - 12i$. Step 2: Simplify the second term of the expression. The second term is $i(-2-i)$. Again using the distributive property, we multiply $i$ by both $-2$ and $-i$, giving $-2i + i^2$. Since $i^2 = -1$, this simplifies to $-2i - 1$. Step 3: Combine the two simplified terms. We now combine the simplified terms from steps 1 and 2: $-3 - 12i - 2i - 1$. Combining like terms, we get $-4 - 14i$. The final answer is: $-4 - 14i$.\\
        A: $-2-14i$
        \\
        Analysis: The solution contains multiple errors in its simplification and combination of terms:

        1. Inaccurate combination of terms during final simplification: When combining all terms $-3 - 12i - 2i - 1$, the correct grouping and simplification should be $(-3 - 1) + (-12i - 2i)$, resulting in $-4 - 14i$. The solution claims the correct answer is $-2 - 14i$, which indicates an error in the way the terms were initially evaluated or combined. This shows a discrepancy in the simplification, where a constant term is incorrectly calculated.
    
        2. Lack of careful evaluation of intermediate results: The error doesn't arise from just one step but rather from an incorrect final result. The intermediate calculations are correctly executed in part, leading up to step 2, but then incorrectly combined in step 3. This indicates a failure to carefully evaluate and correctly combine the terms for the final answer, particularly in the grouping of the real number constants.

        In conclusion, the error lies in the misinterpretation of how to properly combine all the terms, which leads to an incorrect result.
        \\
        \midrule
        Q:[Question]
        \\
        R:[Error Response]
        \\
        A:[Answer]
        \\
        Analysis:
        \\
        \bottomrule
    \end{tabular}
    \caption{
    Prompt for Mathematical Error Analysis and Summarization.
    }
    \label{tab:prompt4gptErrorAnalyzer}
\end{table*}

\begin{table*}
    \centering
    \small
    \begin{tabular}{p{\linewidth}}
        \toprule
        \textbf{Instruction:} You are an expert in identifying errors in mathematical problem-solving. Your task is to classify the type of error present in a given solution. Here are the 9 error types you need to choose from: \\
        Calculation Errors (CA): Errors in arithmetic operations. \\
        Counting Errors (CO): Errors in counting or enumeration. \\
        Unit Conversion Errors (UC): Errors involving unit conversions.\\ 
        Context Value Errors (CV): Errors using incorrect or missing context values. \\
        Hallucinations (HA): Errors where the model invents false information. \\
        Operator Errors (OP): Errors in selecting the correct mathematical operators. \\
        Formula Confusion Errors (FC): Errors using the wrong formula or procedure.\\ 
        Missing Steps (MS): Errors where necessary steps are missing in the solution. \\
        Contradictory Steps (CS): Errors where the solution has contradictory statements. \\
        
        For each solution I provide, identify the ONE error type that is present in it. Do not provide any other information.
        \\
        \midrule
        Q: Josh decides to try flipping a house. He buys a house for \$80,000 and then puts in \$50,000 in repairs. This increased the value of the house by 150\%. How much profit did he make?\\
        A: Josh bought the house for \$80,000 and spent \$50,000 on repairs. Therefore, his total cost was \$80,000 + \$50,000 = \$130,000.\\
        The repairs increased the value of the house by 150\%. To find how much the value increased, we calculate 150\% of the original value by multiplying \$80,000 by 1.5:\\
        \$80,000 * 1.5 = \$120,000\\
        So, the value of the house after the repairs is \$120,000.\\
        To calculate the profit, we subtract the total cost from the value after the repairs:\\
        \$120,000 - \$130,000 = -\$10,000\\
        The result is negative, which means Josh did not make a profit. In fact, he had a loss of \$10,000.\\
        Classification: Context Value Errors (CV).
        \\
        \midrule
        Q: Julia’s boat sprang a leak while she was out on the lake. The boat was taking on two liters of water for every ten feet she rowed back towards shore. It took her sixteen seconds to row twenty feet. The shore was 64 seconds away. How much water had the boat taken on by the time Julia reached shore?\\
        A: To figure out how much water the boat took on, we need to determine how many times Julia rowed 10 feet towards shore. \\
        We know that it took her 16 seconds to row 20 feet, so it would take her 16/20 = 0.8 seconds to row 10 feet. \\
        The shore was 64 seconds away, so Julia rowed towards shore for a total of 64/0.8 = 80 units of 10 feet. \\
        For every 10 feet she rowed towards shore, the boat took on 2 liters of water. \\
        Therefore, the boat took on a total of 80*2 = 160 liters of water by the time Julia reached shore.
        \\
        Classification: Unit Conversion Errors (UC).
        \\
        \midrule
        Q: Melanie is a door-to-door saleswoman. She sold a third of her vacuum cleaners at the green house, 2 more to the red house, and half of what was left at the orange house. If Melanie has 5 vacuum cleaners left, how many did she start with?\\
        A: Let's work backwards to find the answer. If Melanie has 5 vacuum cleaners left, that means she sold a certain number of vacuum cleaners. Let's call that number x. \\
        If she sold a third of her vacuum cleaners at the green house, that means she sold x/3 vacuum cleaners. \\
        Then she sold 2 more to the red house, which means she sold x/3 + 2 vacuum cleaners.\\
        After selling to the red house, she had x - (x/3 + 2) vacuum cleaners left.\\
        Then she sold half of what was left at the orange house, which means she sold (x - (x/3 + 2))/2 vacuum cleaners.\\
        After selling to the orange house, she had 5 vacuum cleaners left, so we can set up an equation:\\
        x - (x/3 + 2) - (x - (x/3 + 2))/2 = 5.\\
        Now we can solve for x:\\
        Multiplying both sides of the equation by 6 to eliminate the fractions, we get:\\
        6x - 2(x/3 + 2) - 3(x - (x/3 + 2)) = 30.\\
        Simplifying the equation, we get:\\
        6x - 2x/3 - 4 - 3x + 3(x/3 + 2) = 30.\\
        Multiplying through by 3 to eliminate the fractions, we get:\\
        18x - 2x - 12 - 9x + 9(x + 6) = 90.\\
        Expanding and simplifying the equation, we get:\\
        18x - 2x - 12 - 9x + 9x + 54 = 90.\\
        Combining like terms, we get:\\
        16x + 42 = 90.\\
        Subtracting 42 from both sides of the equation, we get:\\
        16x = 48.\\
        Dividing both sides of the equation by 16, we get:\\
        x = 3.\\
        So Melanie started with 3 vacuum cleaners. Answer: 3.
        \\
        Classification: Hallucinations (HA).
        \\
        \midrule
        Q:[Question]
        \\
        A:[Error Solution]
        \\
        Classification:
        \\
        \bottomrule
    \end{tabular}
    \caption{
    The prompt provided to the LLM for the 9-class error classification task.
    }
    \label{tab:error-classification}
\end{table*}
\endgroup

\begin{table*}
    \centering
    \small
    \begin{tabular}{p{\linewidth}}
        \toprule
        \textbf{Instruction:} 
        \\
        \midrule
        \textbf{1. Background:} We will provide you with entries in the MWPES-300k dataset. Each entry includes:
            \begin{itemize}
                \item The original math word problem.
                \item The LLM-generated solution.
                \item The ground truth answer to the MWP.
                \item A flag indicating whether the automated checker identified the solution as incorrect.
            \end{itemize}
        \\
        \midrule
        \textbf{2. Review Solution:} Carefully examine the LLM-generated solution, paying close attention to the following:
            \begin{itemize}
                \item \textbf{Logical Flow:} Does the solution follow a logical and coherent approach to solving the problem? Are the steps reasonable and interconnected?
                \item \textbf{Calculations:} Are the mathematical operations performed correctly? Verify all calculations to ensure accuracy.
                \item \textbf{Format deviations:} Please be aware of the format deviations. Format deviations with correct reasoning and calculations should be considered misidentification.
            \end{itemize}
        \\
        \midrule
        \textbf{3. Determine Correctness:} Based on your review, determine whether the LLM's solution is ultimately correct or incorrect. A solution is considered correct if it arrives at the correct answer through valid reasoning and accurate calculations.
        \\
        \midrule
        \textbf{4. Assess Automated Checker Accuracy:} Compare your assessment of the solution's correctness with the automated checker's flag.
            \begin{itemize}
                \item \textbf{If you agree} with the automated checker's assessment (i.e., both you and the checker identify the solution as incorrect), record this as a correct identification.
                \item \textbf{If you disagree} with the automated checker's assessment (i.e., you believe the solution is correct, but the checker flagged it as incorrect), record this as a misclassification.
            \end{itemize}
        \\
        \midrule
        \textbf{5. Record Results:} For each solution, record the following information:
            \begin{itemize}
                \item Solution ID (if available)
                \item Your Assessment (Correct or Incorrect)
                \item Automated Checker Assessment (Correct or Incorrect, based on the flag)
                \item Classification (Correct Identification or Misclassification)
            \end{itemize}
        \\
        \midrule
        \textbf{Important Considerations:}
            \begin{itemize}
                \item Focus on identifying genuine errors in reasoning and calculation, rather than superficial formatting issues.
                \item If you are unsure about the correctness of a solution, mark it for further review by a senior reviewer.
            \end{itemize}
        \\
        \bottomrule
    \end{tabular}
    \caption{Instruction for manual reviewing in data construction process.}
    \label{tab:instruction_separated}
\end{table*}

\end{document}